\documentclass{article}

\usepackage{microtype}
\usepackage{graphicx}
\usepackage{subfigure}
\usepackage{booktabs} 
\usepackage{enumitem}

\PassOptionsToPackage{hyphens}{url}
\usepackage{hyperref}

\usepackage{mdframed}

\usepackage[draft]{icml2024}

\usepackage{amsmath}
\usepackage{amssymb}
\usepackage{stmaryrd}
\usepackage{mathtools}
\usepackage{amsthm}

\usepackage[capitalize,noabbrev]{cleveref}

\theoremstyle{plain}
\newtheorem{theorem}{Theorem}[section]

\theoremstyle{definition}
\newtheorem{definition}[theorem]{Definition}

\theoremstyle{remark}

\theoremstyle{definition}
\newtheorem{example}[theorem]{Example}

\usepackage{csquotes}

\usepackage[colorinlistoftodos]{todonotes}

\icmltitlerunning{Towards Guaranteed Safe AI}

\begin{document}

\twocolumn[
\icmltitle{Towards Guaranteed Safe AI:\\ A Framework for Ensuring Robust and Reliable AI Systems}



\icmlsetsymbol{equal}{*}

\begin{icmlauthorlist}
\icmlauthor{David ``davidad'' Dalrymple}{equal,aria}
\icmlauthor{Joar Skalse}{equal,oxford}
\icmlauthor{Yoshua Bengio}{mila}
\icmlauthor{Stuart Russell}{berkeley}
\icmlauthor{Max Tegmark}{MIT}
\icmlauthor{Sanjit Seshia}{berkeley}
\icmlauthor{Steve Omohundro}{beneficial}
\icmlauthor{Christian Szegedy}{xai}
\icmlauthor{Ben Goldhaber}{far}
\icmlauthor{Nora Ammann}{pibbss}
\icmlauthor{Alessandro Abate}{oxford}
\icmlauthor{Joseph Y. Halpern}{cornell}
\icmlauthor{Clark Barrett}{stanford}
\icmlauthor{Ding Zhao}{cmu}
\icmlauthor{Tan Zhi-Xuan}{MIT}
\icmlauthor{Jeannette Wing}{columbia}
\icmlauthor{Joshua B. Tenenbaum}{MIT}
\end{icmlauthorlist}

\icmlaffiliation{aria}{UK Advanced Research + Invention Agency}
\icmlaffiliation{mila}{Mila - Quebec AI Institute / U. Montreal}
\icmlaffiliation{berkeley}{UC Berkeley}
\icmlaffiliation{MIT}{Massachusetts Institute of Technology}
\icmlaffiliation{beneficial}{Beneficial AI Research}
\icmlaffiliation{xai}{x.AI}
\icmlaffiliation{oxford}{Oxford University}
\icmlaffiliation{far}{FAR AI, Inc.}
\icmlaffiliation{pibbss}{Principles of Intelligent Behaviour in Biological and Social Systems}
\icmlaffiliation{cornell}{Cornell University}
\icmlaffiliation{stanford}{Stanford University}
\icmlaffiliation{cmu}{Carnegie Mellon University}
\icmlaffiliation{columbia}{Columbia University}

\icmlcorrespondingauthor{David ``davidad'' Dalrymple}{david.dalrymple@aria.org.uk}
\icmlcorrespondingauthor{Joar Skalse}{joar.skalse@cs.ox.ac.uk}

\icmlkeywords{Machine Learning, ICML}

\vskip 0.3in
]



\printAffiliationsAndNotice{\icmlEqualContribution} 

\begin{abstract}
Ensuring that AI systems reliably and robustly avoid harmful or dangerous behaviours is a crucial challenge, especially for AI systems with a high degree of autonomy and general intelligence, or systems used in safety-critical contexts. In this position paper, we will introduce and define a family of approaches to AI safety, which we will refer to as \emph{guaranteed safe} (GS) AI. The core feature of these approaches is that they aim to produce AI systems which are equipped with \emph{high-assurance quantitative safety guarantees}. This is achieved by the interplay of three core components: a \emph{world model} (which provides a mathematical description of how the AI system affects the outside world in a way that appropriately handles  both Bayesian and Knighting uncertainty), a \emph{safety specification} (which is a mathematical description of what effects are acceptable), and a \emph{verifier} 
(which provides an auditable proof certificate that the AI satisfies the safety specification relative to the world model). We outline a number of approaches for creating each of these three core components, describe the main technical challenges, and suggest a number of potential solutions to them. We also argue for the necessity of this approach to AI safety, and for the inadequacy of the main alternative approaches. 
\end{abstract}

\section{Introduction}\label{sec:introduction}

We introduce and define a family of approaches to AI safety, collectively referred to as \emph{guaranteed safe} (GS) AI. These approaches aim to provide high-assurance quantitative guarantees about the safety of an AI system’s behaviour through the use of three core components –- a formal \emph{safety specification}, a \emph{world model}, and a \emph{verifier}. We will argue that this strategy is both promising and underexplored, and contrast it with other ongoing efforts in AI safety. We will also outline several ongoing avenues of research within the broader GS research agenda, identify some of their core difficulties, and discuss approaches for overcoming these difficulties. Central examples of agendas which fall under the GS AI family include \citet{szegedy2020promising, wing2021, seshia-cacm22a, russell2022, tegmark2023provably, dalrymple2024,bengio2024cautious}.

Critical infrastructure and safety-critical systems are required to comply with high safety standards. For example, aircraft, nuclear power plants, and medical devices are subject to exceptionally rigorous safety certification. Moreover, it is plausible that there will soon be AI systems that are at least as safety-critical as these systems (e.g.\ as AI systems are increasingly deployed in safety-critical contexts, or given greater capabilities and autonomy), which means that they should be required to adhere to standards of safety that are at least as strict. Above some risk or capability thresholds, the burden of demonstrating such safety guarantees should be on the systems' developers, and the provided evidence must be adequate to justify high confidence that the AI system is safe. We will argue that approaches based only on experimental tests are insufficient for producing such safety guarantees. We will also argue that GS AI presents research avenues that could plausibly produce such safety guarantees in a satisfactory and feasible manner, while remaining competitive with AI systems without such guarantees.

In Section~\ref{sec:background}, we provide the general background and context for this paper, including a brief overview of the AI safety problem, and some avenues for producing quantitative safety assessments through the use of empirical techniques. In Section~\ref{sec:GS_AI}, we introduce and define GS AI, together with an extensive discussion of the spectrum of approaches that fall under this agenda, the main challenges to these approaches, and some potential solutions. We see the different approaches as comprising a portfolio of complementary R\&D efforts.
In Section~\ref{sec:examples}, we offer several examples of concrete practical problems, and how they may be solved with a GS AI approach.
In Section~\ref{sec:discussion} we provide further high-level discussion of our proposals, including their feasibility and benefits. In Section~\ref{sec:existing_work} we list some related work, and in Section~\ref{sec:conclusion} we conclude the paper.

\section{Background}\label{sec:background}

In this section, we provide the background that is required to understand the rest of this paper and its context. This includes an overview of the AI safety problem, how to classify AI systems based on the level of risk they pose, and a discussion of   empirical safety assessments. 

\subsection{The AI Safety Problem}\label{sec:AI_safety}

A number of prominent AI experts have argued that sufficiently advanced AI systems may threaten the survival of the human species, or lead to our permanent disempowerment, especially in the case of AI systems that are more intelligent than humans. Such concerns have been raised by \citet{butler1863}, \citet{asimov1942}, \citet{wiener1950}, \citet{turing1951}, \citet{good1959}, \citet{minsky1984}, \citet{moravec1988}, \citet{vinge1993}, \citet{joy2000}, \citet{yudkowsky2001}, \citet{bostrom2002}; as well as, more recently, \citet{russell2019human, testimony_russell}, \citet{tegmark2018life}, \citet{bostrom2014superintelligence}, \citet{interview_pearl}, \citet{testimony_bengio}, \citet{interview_hinton}, \citet{testimony_amodei}, and others. These experts have provided many different arguments in support of these concerns, which we will not reproduce here in full. However, here are a few very brief summaries of some of their central arguments:

\begin{enumerate}
    \item For an AI system to reliably solve a complex problem in an open-ended domain where i.i.d. assumptions are not valid, we must provide it with a formalisation of what it means to solve that problem. However, it is difficult to create such specifications without leaving a substantial gap between what was specified and what was intended. This issue has been observed empirically in current AI systems \citep[e.g., ][]{krakovna_specification_2020, reward_misspecification, pang_reward_2023}, and studied theoretically \citep[e.g., ][]{subset_features, skalse_defining_2022, skalse2022invariance, karwowski2023goodharts, skalse2024starc}. This suggests that it is difficult to engineer AI systems to robustly act in accordance with our intentions, especially in unprecedented situations (which may also arise as AI itself increasingly reshapes its contexts of use). 
    \item In a conflict of interests, greater intelligence is a substantial advantage. For example, in a chess game between a novice and a grandmaster, we should expect the grandmaster to win. More generally, the reason why humanity is the most powerful species on the planet is primarily that we have the greatest capacity to devise and carry out complex plans for reshaping our environment. Technological innovation is also the greatest driver of economic growth and military capabilities. 
    Thus, if there are ever AI systems that are substantially more 
    intelligent
    than humans, and which are not aligned with human interests, then we should expect human interests to be marginalised.
    \item Even if the goals of an AI system are specified correctly, it may still fail to internalise these goals in the intended way. For example, one way to maximise a reward signal that is provided by human feedback may not be to do what the humans wish, but rather to take control of the reward mechanism \citep{cohen2022advanced}. Similar phenomena have been observed empirically in current AI systems \citep{shah2022goal,langosco2023goal}, and studied theoretically \citep{hubinger2021risks}.
\end{enumerate}

Existing attempts to solve these problems have so far not yielded convincing solutions, despite rather extensive investigations \citep{ji2024ai}. This suggests that the problem is fundamentally hard, on a technical level.
For a more complete and in-depth treatment of the arguments for why future AI may pose an existential risk to humanity, see e.g.\ \citet{bostrom2014superintelligence, russell2019human}.

Other AI experts have also pointed to more immediate risks from AI systems. For example, generative AI may facilitate the spread of disinformation, by enabling the creation of convincing deepfakes or by making it cheaper and easier to produce large volumes of content \citep{brundage2018malicious}. AI decision making may also unfairly and systematically disadvantage certain groups, even when unintended by the system's creators \citep[e.g.,][]{kleinberg2016inherent, buolamwini2018gender, fairness_review_1, fairness_review_2}. Recommender systems may facilitate invasions of privacy and the spread of extreme content \cite{stray2021designing, carroll2022estimating, NBERw26669, Settle_2018, hostileaudience}. AI may also enable large-scale surveillance \cite{feldstein2019global} or the centralisation of economic or political power \citep{crawford2021atlas,brynjolfsson2023big}. For an overview of some of these issues, see, e.g., \citet{FATEoverview, hendrycks2023overview, brundage2018malicious}.

These two perspectives on the risks from AI are not mutually exclusive. Moreover, they both point to similar sets of technical challenges. The \emph{AI safety problem} is the problem of ensuring that AI systems reliably and robustly act in ways that are not harmful or dangerous, including (but not limited to) cases where those AI systems are more intelligent than humans. In this paper, we are proposing a family of engineering strategies for solving this problem in general.

Note that the problem of ensuring that AI is not harmful to humans comprises both a technical and a societal problem; solving the technical problems is not sufficient if the solutions are not globally implemented. In this paper, we will primarily focus on the technical aspect of the AI safety problem. For an overview of some of the political and sociological challenges, see e.g.\ \citet{lazar2023ai,bostrom2014superintelligence, alaga2023coordinated, koessler2023risk, sastry2024computing, schuett2023best}.

\subsection{AI Safety Levels}\label{sec:AI_safety_levels}

The levels of precaution that are appropriate for a given AI system depend on the capabilities of that system. To classify the relevant levels of capability, Anthropic has introduced a framework that they call AI Safety Levels (ASL)\footnote{The ASL framework can be understood as a type of \emph{Safety Integrity Level} (SIL) framework, which are also the basis of standards in various other industries such as MIL-STD-882, IEC 61508, ISO 26262, etc. The ASLs are loosely modelled after the US government’s biosafety level (BSL) in particular.} as part of their voluntary safety commitments \citep{anthropic_responsible_scaling}. This framework classifies AI systems into the following high-level categories: 

\begin{enumerate}
    \item ASL-1 refers to systems which pose no meaningful catastrophic risk.
    \item ASL-2 refers to systems that show early signs of dangerous capabilities (such as the ability to give instructions on how to build biological weapons) but whose risks still appear importantly bounded (e.g.~due to the information not being practically useful, not reliable, or offering no significant improvement over what is achievable relative to non-AI baselines such as search engines or textbooks). Many current language models appear to be ASL-2.
    \item ASL-3 refers to systems that substantially increase the risk of catastrophic misuse compared to non-AI baselines, or that show low-level autonomous capabilities.
    \item ASL-4 and higher (ASL-5+) are not yet defined in as much detail, but generally refer to human level and superhuman levels of intelligence, and thus qualitative escalations in the potential for catastrophic misuse and autonomy.
\end{enumerate}

We refer to this classification scheme throughout the paper. Note that similar classification schemes also have been defined by other authors \citep[e.g.,][]{fli_safety_levels, khlaaf2022hazard}.  

\subsection{Empirical Safety Assessments}\label{sec:probabilistic_safety_assessment}

Current approaches to validating frontier AI models prior to deployment lean on independent testing and red-teaming \citep[as e.g.\ ][]{ziegler2022adversarial, perez2022red, ganguli2022red, ouyang2022training,  khlaaf2023toward}. Such methods can find concrete examples of unsafe behaviour, which can then be rectified by the developer or lead to the decision to halt training and deployment.
However, these testing regimes do not provide a rigorous, quantifiable safety guarantee: red-teamers could fail to find serious failures, while a model still harbours such failure modes. In the extreme case, a model could have been backdoored in a way that is cryptographically hard to detect without knowing the trigger \citep{guo2021overview}. Even in the absence of such malfeasance, weaknesses in AI systems can remain undetected even after extensive testing. 
For example, ChatGPT was evaluated in great detail over a period of several months \citep{openai2024gpt4}, and yet users found ways to circumnavigate its safety precautions within just a single day of it going public \citep{jailbreaking}. This issue is likely to be even more pertinent for more capable systems tasked with solving more difficult problems. Moreover, it is also important to note that AI systems often will be deployed in \emph{adversarial} settings, where human actors (or other AIs) actively try to break their safety measures. In such settings empirical evaluations are likely to be inadequate; there is always a risk that an adversary could be more competent at finding dangerous inputs, unless you have a strong guarantee to the contrary.

We argue that to obtain a high degree of confidence in a system we need a positive safety case providing quantifiable guarantees, using either or both empirical and theoretical arguments (also see \citet{khlaaf2019assuringMLauto, khlaaf2022hazard} for a similar argument). This is not unique to AI. About 25\% of bridges built in the 1870s collapsed within the decade \citep{mccullough2001great}, before a deeper theoretical understanding of civil engineering reduced bridges' failure rates to less than 0.4\% per decade \citep{Cook2014BridgeFR}. When quantitative safety assessments are required from developers, this makes it possible for society to mandate a clear level of safety, in terms of, e.g., the frequency per year of adverse events of certain magnitudes (this is called a societal risk curve). At its simplest, this could consist of extensive testing: if an autonomous vehicle drives a million miles safely without intervention of a test driver, then we can conclude that the failure rate probably is less than one in a million miles when deployed in the same operating domain. However, note that such inferences may only be valid within the bounds of unreasonable assumptions. For example, if human drivers start driving more aggressively around autonomous vehicles after they are more used to them \citep{liu2020bully}, then this safety assessment might cease to hold. 

Stronger empirical bounds (i.e.\ with weaker assumptions) may be obtained through methods such as adversarial testing, and/or testing in simulations with domain randomisation. Even stronger bounds may be obtained through a more mechanistic understanding of the system. Fault trees \citep{nieuwhof1975introduction} are a common safety engineering technique that allows for quantitative analysis, which can be interpreted as deduction in probabilistic logic. For example, if two redundant components can be shown to have a failure rate of $\leq n^{-1}$ and the failure rates are independent, the combined failure rate would be $\leq n^{-2}$. This could occur for example in an autonomous vehicle case when performing object detection on different sensors (e.g.\ LIDAR, vision, radar), or in a generative model case when using an ensemble of different models to detect malicious inputs. A similar approach might be applicable even in the case of a single monolithic model, by leveraging approaches like mechanistic interpretability that seek to decompose internal representations of the model \citep{zhang2021interpretability, gao2023interpretability, bricken2023towards, michaud2024opening}.

An alternative approach to obtaining stronger bounds relies on theoretical understanding of the system. For example, a rigorous theory for how deep networks generalise \citep[as built towards by, e.g., ][]{introtoCLT, Watanabe_2009, Watanabe_2018, SGDBayes, NNentropy} might enable principled extrapolation from empirical testing on a limited validation domain to a broader test domain. Combined, these approaches could enable carefully conducted empirical evaluations to provide substantially stronger safety bounds than exist for contemporary frontier models.

However, any empirical evaluation must ultimately rely on some relatively strong assumptions, such as the distribution of inputs used to validate the models being sufficiently similar to those they are deployed on. This makes it challenging for an empirical approach to rule out instances of deceptive alignment, where a system is acting to subvert the evaluation procedure by detecting features of the input distribution that are unique to the test environment \citep{hubinger2021risks}. It also makes it challenging to give long-horizon safety guarantees, where the distribution of inputs is likely to naturally shift over time.\footnote{Note that the usage of powerful AI is likely to itself create situations that are novel and unprecedented, which makes this a point of particular importance. Stated differently, we should \emph{assume} that \enquote{distributional shift} will occur.} To achieve stronger safety guarantees, which will become increasingly critical with ASL-3 and beyond, we therefore expect it to be necessary to use a {\em model-based} approach. We discuss this approach in the following section.

\section{Guaranteed Safe AI}\label{sec:GS_AI}

In this section, we will introduce and characterise a family of approaches to the AI safety problem, which we refer to as guaranteed safe (GS) AI. We will first provide a definition of GS AI, together with a high-level overview. We will then discuss each of the core components of GS AI, namely a \emph{world model}, a \emph{safety specification}, and a \emph{verifier}. For each of the core components, we describe its role in the overall architecture towards providing high-assurance safety guarantees, highlight some of the key challenges in trying to implement these components, and discuss current approaches to overcoming those challenges.

\subsection{Definition of GS AI}\label{sec:defining_SG_AI}

The core feature of the GS approach to AI safety is to produce systems consisting of an AI agent and other physical, hardware, and software components which together are equipped with a high-assurance quantitative safety guarantee, taking into account bounded computational resources.  This can be contrasted against approaches to AI safety which primarily rely on empirical evaluations, or informal arguments based on qualitative or pre-theoretic intuitions.

A high-assurance quantitative safety guarantee may take the form of a formal proof that the system always will adhere to some safety specification\footnote{In this paper, we use the term \enquote{safety} not in the strict sense used in formal methods~\cite{AlpernSchneider87} but in the broader sense used in AI for the most important specifications of correct behaviour.} (or distribution over safety specifications) for all inputs, relative to a model (or a distribution over models) of world dynamics. Alternatively, it may be a reliable and sound upper bound on the probability of violating a safety specification. However, especially when we cannot find a proof certificate, or cannot formally define the desirable or undesirable behaviour, it may also take the form of an estimate of an upper bound on the probability of harm, with the estimate asymptotically converging with computational resources in order to guarantee a constraint on desirable behaviour. This gives us the following definition (for a more formal definition, see Appendix~\ref{appendix:formal_definition}):

\begin{definition}\label{def:GSAI}
A Guaranteed Safe AI system is one that is equipped with a quantitative safety guarantee that is produced by a (single, set of, or distribution of) \emph{world model(s)}, a (single, set of, or distribution of) \emph{safety specification(s)}, and a \emph{verifier},
satisfying the following criteria:
\begin{enumerate}
    \item The probabilistic specification encodes societal risk criteria, which should ideally be determined by collective deliberation.
    \item The verifier provides a quantitative guarantee (in the form of a proof certificate, probabilistic bound, asymptotic guarantee, or other comparable assurance) that the AI system satisfies the specification with respect to the world model.
    \item All potential future effects of the AI system and its environment  relevant to the safety specification should be modelled (and -- if feasible -- conservatively over-approximated) by the world model.
\end{enumerate}
\end{definition}

\begin{figure*}
    \centering
    \includegraphics[width=0.8\textwidth]{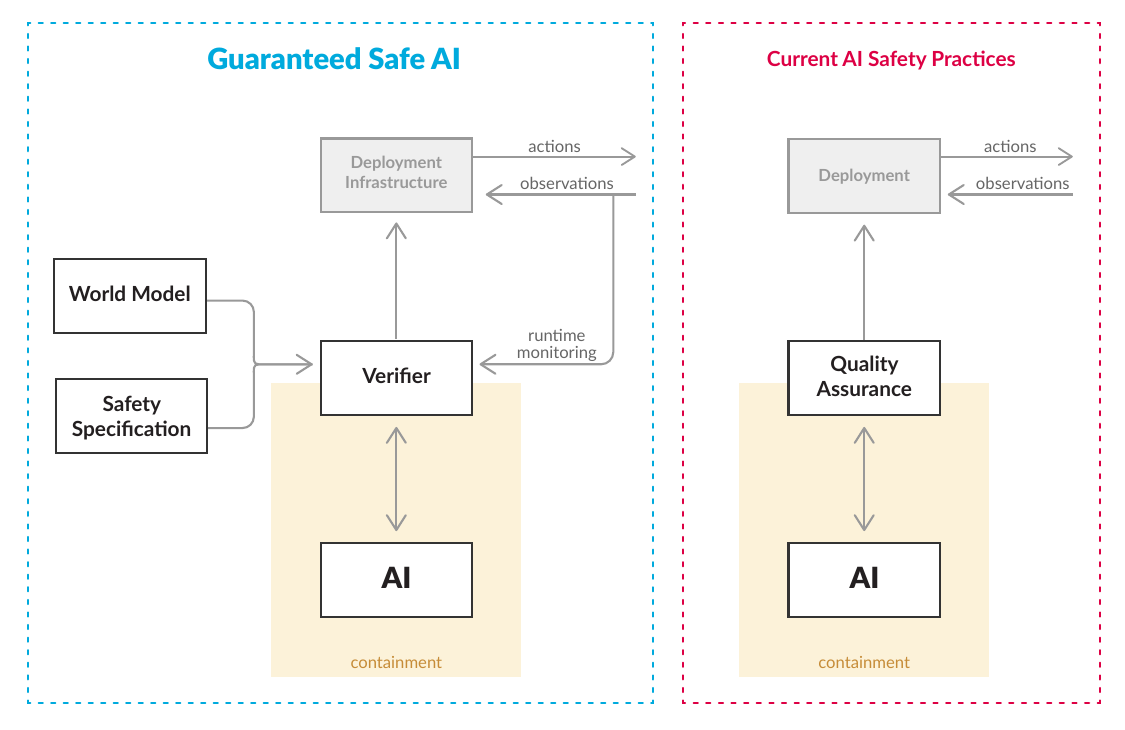}
    \caption{The GS AI approach builds on three components, namely a \emph{world model} that describes the environment of the AI system, a \emph{safety specification} that describes desirable safety properties and is expressed in terms of the world model, and a \emph{verifier} that provides a quantitative guarantee of the extent to which an AI system satisfies the safety specification. In contrast, current AI Safety practices rely primarily on quality assurance (e.g.~evaluations) to decide if an AI system is safe, which is insufficient for safety critical applications.}
    \label{fig:GS-AI-overview-fig}
\end{figure*}

There is much to unpack in this definition. Before moving on, we will therefore provide a brief explanatory example of what each of the core components of Definition~\ref{def:GSAI} could look like, together with a motivation for their necessity. Later in this section, we will provide a more in-depth discussion of what each of these components may look like, what challenges they come with, and how those challenges may be overcome. An overview is also provided in Figure~\ref{fig:GS-AI-overview-fig}. But first, let us provide some intuition. 

A safety specification corresponds to a property that we wish an AI system to satisfy. For example, we may wish that an AI system never takes any actions that may plausibly cause a human to be harmed. If we have a formal definition of harm, as well as a formal definition of causation, then this safety property could be turned into a well-defined formal specification.  Of course, neither of these terms are easy to formalise in general, but theoretical proposals exist \citep[e.g.\ ][]{beckers2022quantifying, harm2022, Halpern16, Pearl09}. It is sometimes also possible to operationalise harm in a context-specific manner \citep{khlaaf2023toward}. Other safety specifications may also be desirable. For example, we may wish to require that an AI system is \enquote{truthful}, or that it can offer \enquote{explanations} for its actions, etc.
In Section~\ref{sec:safety_spec}, we will provide a more detailed overview of some possible methods for obtaining safety specifications, as well as their advantages and challenges. Note that due to  difficulties with formulating adequate safety specifications, it may be desirable to use multiple safety specifications, or to use \emph{distributions} of safety specifications, obtained through some learning process (or otherwise).

Next, many desirable safety specifications necessarily require a \emph{world model} (or distribution over world models) that describes the dynamics of the AI system's environment. For example, suppose we want to ensure that an AI system never takes any actions that leads a human to be harmed, according to some (possibly ambiguous) definition(s) of \enquote{harm}. In order to do this, we need a model that describes whether a given action is likely to lead to a human being harmed (in a given context). More generally, without a world model we can only verify specifications defined over input-output relations, but it is often desirable to instead verify specifications over input-\emph{outcome} relations. We may also want to define predicates such as \enquote{harm} in terms of not only directly observable quantities (like a human-provided label), but also in terms of unobserved or even counterfactual variables (such as how a group of wise humans would hypothetically judge an outcome). Of course, we must also be able to trust the correctness of this world model, which means that it ideally should be interpretable and understandable. For more discussion of how to create such a world model (or distribution of world models), see Section~\ref{sec:world_model}. 
Note that the world model need not be a \enquote{complete} model of the world; the level of detail and abstraction that is adequate depends on the safety specification and the AI system's context of use.

Given a safety specification and a world model, we also need a way to produce quantitative assurances for a given AI system. In the most straightforward form, this could take the shape of a formal proof that the AI system (or its output) satisfies the safety specification relative to the world model. This is akin to traditional formal verification \citep[see e.g.\ ][]{Baier2008, leino2023program, seligman2023formal}. Of course, such formal verification is often hard to produce, even for relatively simple computer programs, and traditional formal methods face unique challenges for AI systems~\cite{seshia-arxiv16}. However, further progress in automated reasoning and theorem proving due to integration with data-driven learning \citep[see e.g.\ ][]{lample2022hypertree,seshia-pieee15,wu2022autoformalization,first2023baldur,Trinh2024} could make this substantially easier, and might also scale with further progress in AI more generally. Moreover, if a direct formal proof cannot be obtained, there are weaker alternatives that would still produce a quantitative guarantee. For example, it may take the form of a proof that bounds the probability of failing to satisfy the safety specification, or a proof that the AI system will converge towards satisfying the safety specification (with increasing amounts of data or computational resources, for example). Indeed, many model-based AI algorithms have been designed to satisfy exactly these sorts of guarantees \cite{McMahan2005, Junges2016, mathews2022finite, Hasanbeig2023-HASCRL}. For a more detailed discussion, see Section~\ref{sec:verifier}.

Note that the GS AI approach remains agnostic about both how the \enquote{core} AI system was produced, and about what containment or boxing methods \cite{armstrong2012thinking} are used. If the verifier and the world model together can establish that the AI (or the AI's output) satisfies the safety specification, a quantitative safety guarantee is obtained (even if the AI system remains uninterpretable, etc).

\subsection{The World Model}\label{sec:world_model}

In this section we will discuss the world model, including both a more detailed discussion of what counts as a world model, as well as some possible strategies for constructing such a world model (including their challenges and benefits).

The world model needs to answer queries about what would happen in the world as a result of a given output from the AI. It must also describe the state of the world at a level of granularity that is sufficient for expressing the safety specifications we are interested in.\footnote{Note that a world model with a more abstract state-space may make it easier to express certain safety specifications, or make verification more tractable, but that this may come at the cost of making the predictions less accurate. Also note that the world model need not make predictions about arbitrary properties of the world, and even about properties on which they do make predictions, these predictions need not necessarily be precise.} A key challenge -- which we will discuss more below -- is finding a satisfactory solution to handling both Bayesian and Knightian uncertainty. World models also serve to elucidate the AI system designers’ assumptions, and we must be mindful that those assumptions may hold only part of the time. Hence, the domain of applicability and epistemic uncertainty regarding different pieces of the world model must be represented and taken into account. For these reasons, the world model, or relevant aspects of it, should be auditable or monitorable at run time.

\begin{figure*}
    \centering
    \includegraphics[width=1\linewidth]{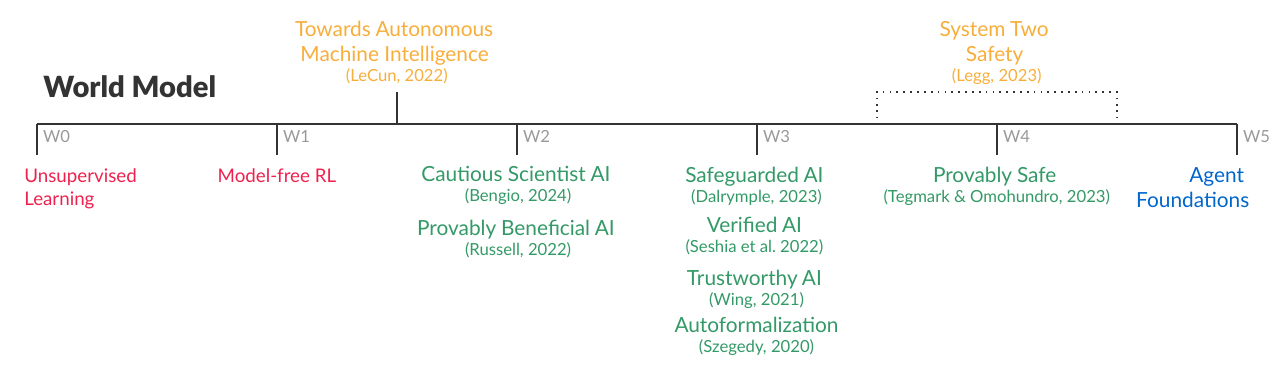}
    \caption{Different approaches for building world models can be projected onto a spectrum according to how much safety they would grant if successfully implemented. Approaches listed in green fall into the GS AI family. Approaches in yellow may qualify (depending on underspecified aspects of the approach) or come close to qualifying as GS AI. In our view, approaches in red fail to provide high-assurance quantitative safety guarantees, and thus do not qualify as GS AI.}
    \label{fig:model_spectrum}
\end{figure*}

There are many possible strategies for creating world models. These strategies can roughly be placed on a spectrum, depending on how much safety they would grant if successfully implemented (see also Figure~\ref{fig:model_spectrum}):

\begin{itemize}
    \item Level 0: You have no world model. Instead, assumptions about the world are implicit in the training data and in aspects of the implementation of the AI system.
    \item Level 1: You use a trained black-box world simulator as your world model.
    \item Level 2: You use a machine-learned generative model of probabilistic causal models, which you can test  by checking whether it assigns sufficient credence to specific human-made models (such as e.g.\ models proposed in the scientific literature).
    \item Level 3: You use (a distribution over) probabilistic causal model(s), potentially generated with the help of machine learning, that are fully audited by human domain experts.
    \item Level 4: You use world models about real world phenomena that are formally verified as sound abstractions of fundamental physical laws.
    \footnote{By \enquote{sound abstraction} we mean the concept which has also been called \enquote{reverse simulation} or \enquote{oplax coalgebra morphism}. Formally, a sound abstraction (of a dynamics $f : X\to \Delta X$ on a state space X) is an abstracted state space A, an abstraction relation $a : X \to \mathcal{P}A$, and a dynamics $g : A \to \Delta A$ such that $f\fatsemi a \preceq a\fatsemi g$.}
    \item Level 5: You have no world model, and instead use safety specifications universally quantified over the entire set of all possible worlds.
\end{itemize}

To get a better intuition for what these levels may look like, let us discuss a number of potential approaches to constructing world models. First of all, in some cases, it may be feasible for engineers to manually create an adequate world model. This has been done in settings where the operating environment is known or controllable, for e.g.\ controllers in airplanes \citep{autopilot_verification,fremont-cav20} and self-driving cars \citep{car_verification_case_study, fremont-itsc20}. 
Such models are also commonly used in scientific research, including epidemic simulators \cite{Broeck2011} and particle physics simulators \cite{etalumis}. 
For AI systems, probabilistic programs have been shown to be a promising formalism for world modeling \citep[e.g., ][]{blog,matheos2020transforming,gothoskar20213dp3,scenic-mlj22}. 
In principle, this manual approach is likely to provide the best understanding of the assumptions underlying the world model, as well as the delineation of its domain of applicability. This approach would produce a world model on Level 4 or 5. However, for AI systems that directly interface with very complex systems (such as e.g.\ human users, the world economy, or sensitive ecosystems), it may not be possible to create sufficiently accurate world models in a fully manual way.

In such cases, the world model must instead be machine learned (or automatically generated by some other means). One possible approach to creating a world model with AI is to use large language models (or similar systems) to write probabilistic programs that correspond to the system(s) being modelled \citep[e.g.,\ ][]{wong2023word,wong2023learning,elmaaroufi-arxiv24,tang2024worldcoder, lew2020thoughtppl}. This method has the potential to be scalable, since most of the work is offloaded to AI systems. This method may also produce world models that are interpretable by default  \citep{grand2023lilo}, if the language models are trained on code written by human programmers, or if the language used to generate pieces of the world model is forced or encouraged to be interpretable through probabilistic translation to and from natural language \citep{jacob2023consensus}. If successful, this approach could produce a world model on Level 2, 3 or possibly 4,  in the classification above (depending on the extent to which the resulting model is audited). One of the main challenges with this approach would likely be to ensure that the world model has a high predictive accuracy. 

Another approach is to learn a world model from data. However, this approach comes with a number of challenges. In particular, many machine learning methods are prone to being confidently incorrect in novel situations. On a theoretical level, this problem can be solved by Bayesian induction, which infers a distribution over world models (i.e. \emph{theories}) from data \citep{kemp2010probabilistic}. Unfortunately, exact Bayesian induction is typically not computationally tractable \citep{hutter2003gentle}. However, recent advances in both Bayesian deep learning \citep{hollmann2023tabpfn, hu2023gflownetem} and  probabilistic programming suggest that \emph{approximate} Bayesian induction can be made tractable \citep[e.g.\ ][]{Baydin2019}. Using deep learning to parameterize generative flow networks \citep{hu2023gflownetem}, one can train large neural networks to implicitly estimate and sample from the posterior distribution $P(T \mid D)$ over theories $T$ given data $D$, including cases where $T$ itself represents a structured (causal) model \citep{deleu2022bayesian, ke2022learning, deleu2023joint}. Using probabilistic programming, one can leverage sequential Monte Carlo (SMC) methods to gradually infer a theory $T$ (represented as probabilistic program) as new data is observed \citep{saad2023sequential}, or by incrementally refining an initially coarse 3D world model  \citep{gothoskar2023bayes3d,stuhlmuller2015coarse}. Both of these approaches can potentially be combined\footnote{For example, the trained neural network can be used to guide SMC sampling \citep{zhao2024probabilistic} and the resulting samples can in turn be used as training data for the neural network.}, and also benefit from more computation: by increasing the size or training time of the neural network used to approximate $P(T \mid D)$, convergence to the true posterior is expected. This means that we can continue training at run-time, or at least estimate the error made by the neural network through a sampling process. Similarly, by increasing the number of parallel samples, sequential Monte Carlo (SMC) methods asymptotically converge \citep{del2006sequential}. We can further decrease the risks associated with insufficient training or sampling by encouraging the inferred theories to be human-inspectable (e.g.~by conditioning on the constraint that theories can be converted to natural language and back with minimal error). If successful, this approach would produce a world model on Level 2 (or 4, if the model passes additional checks). See \citet{bengio2024cautious} for a more complete discussion of this approach. 

A very ambitious approach would be to use world models that are formally verified as being sound abstractions of the basic laws of physics. Note that while physics is not a completed field, there is good reason to believe that our current best theories are completely accurate in certain domains: for example, quantum gravity effects appear irrelevant to contemporary AI systems \citep[see e.g.\ ][]{carroll2021quantum}. 
We can therefore be confident that a system truly satisfies a specification if that specification is verified for that system relative to our best theories of physics, and that specification also includes the requirement that the system is not moved beyond the domains where our theories are known to be accurate. \footnote{It is also worth acknowledging that even a perfect model of physical dynamics is insufficient for safety, since safety-critical queries (e.g.\ whether a given molecule is toxic to humans) presumably will depend on facts about the initial conditions (e.g.\ of human cells) that are not deducible from physics alone. This must be addressed by inference about initial conditions and boundary conditions from data and observations, tempered by appropriately conservative epistemic frameworks incorporating Bayesian and Knightian uncertainty.} If successful, this would produce a world model on Level 4. One of the main challenges for this approach will be the immense computational complexity of producing and verifying such models. To reduce the computational complexity of this approach, approximate world models can be combined as compositional modules of one another, akin to how science approximates fluids as made of molecules made of atoms made of elementary particles, etc.
For more details, see \citet{tegmark2023provably}.

Note that in many cases, it will likely be necessary for the world model to model the behaviour of humans. Moreover, humans are very complex, and it seems dubious to presuppose that it is possible to create a model of human behaviour that is both interpretable and highly accurate (especially noting that such a model itself would constitute an AGI system). There may thus in some cases be a fundamental trade-off between interpretability and predictive accuracy. However, note that interpretability can be maintained if human behaviour is modelled using \emph{nondeterminism}.\footnote{Here \enquote{nondeterminism} should be understood in the same sense as in \enquote{nondeterministic automata} and \enquote{nondeterministic Turing machines}, rather than as a synonym for \enquote{probabilistic}. In other words, the world model may specify a \emph{set} or \emph{class} of possible (potentially stochastic) behaviours for each human, and require the verifier to validate the safety specification for each behaviour profile in this set or class.} Existing work on incorporating non-determinism into mathematical modelling comes form mathematics \citep{mio21combiningnondet, halpern2013, kosoy2021infrabayes, liellcock2024}, computer science \citep{brookes1984comseq, brookes1984failures, thacker2006VV, zhang2022viper} as well as robotics \citet{lavalle1996}. The same applies to other highly complex systems. Also note that the levels in the classification above can be mixed within a single GS AI system. For example, a system can use stronger models of its engineering systems and weaker models of its interactions with the social domain (e.g., you may want to model your sensors with Level 5, but you will not be able to use that rigour for e.g.\ social phenomena). 

A potential argument against the GS research agenda is that the world may be so complex that it is infeasible even in principle to create a sufficiently accurate world model (because of chaotic dynamics, etc). We have several responses to this point. First of all, a world model can (and should) of course include model uncertainty, and this uncertainty can be taken into account when the safety specifications are verified.\footnote{As a very simple example, suppose that the safety specification is given relative to a finite time horizon of $n$ steps, and that we have reason to believe that the world model is wrong with probability at most $\epsilon$ per step over the first $n$ steps. Then if a policy can be proven to satisfy this specification relative to the world model, we should believe that it will satisfy the specification with probability at least $(1-\epsilon)^n$ in the real world.} In this way, the strength of the resulting formal guarantees will be appropriately sensitive to the reliability of the world model. Moreover, many of the most concerning loss-of-control scenarios with advanced AI systems involve cases where the AI is assumed to be able to generate and execute complicated plans with high reliability. In other words, it is reasonable to assume that if AI systems can be powerful enough to pose a serious danger to humanity, then it is possible to create sufficiently accurate world models.  Finally, we would like to note that this argument essentially is fully general. \emph{Any} strategy for creating safe AI systems must rely on some beliefs or assumptions about the real world, and these assumptions could be wrong; a world model simply makes these assumptions explicit. Stated differently, if it is impossible to create world models that are sufficiently accurate to ensure that an AI system adheres to some specification, then it is presumably in general impossible to ensure that this specification is satisfied by that system.

\subsection{The Safety Specification}\label{sec:safety_spec}

In this section we will discuss the safety specification(s), the difficulties with creating such specifications, and some potential strategies for overcoming these difficulties.

Much of the AI safety literature formalises desirable system behavior in terms of reward functions \citep[e.g.\ ][]{amodei2016concrete}. A safety specification is in general different from a reward function, though they include bounded reward functions as a special case. In particular, a safety specification may include properties defined by probabilistic temporal logics, causal counterfactual queries, or even hyperproperties, which reward functions cannot typically express \citep{seshia-atva18, pmlr-v216-skalse23a, subramani2024expressivity}. However, safety specifications cannot include unbounded evaluations of the world-state.\footnote{The reason for this requirement is that it should be possible to completely satisfy the specification in order to relieve optimisation pressure. In order for it to be possible, it must at least be conceivable, i.e., the specification’s level of satisfaction needs to be bounded at least in theory.}
Safety specifications could be expressed in terms of probabilistic computation tree logic (PCTL) \citep{PCTL}, or in terms of conjunction of linear inequalities of reachability probabilities, for example. Specification logics can also invoke neural components as predicates \citep{mao2019neuro}, allowing us to learn the semantics of concepts that are hard to manually specify.

The appropriate safety specification depends on the application domain and context \citep{khlaaf2023toward}. Task-specific AI services often afford contextual operationalisations of safety, while more general-purpose AI assistants may require specifications that invoke broad definitions of harm \citep{beckers2022quantifying,london2023beneficent}.\footnote{See Appendix~\ref{appendix:harm} for a short discussion of a potential definition of harm along the lines of \citet{harm2022}.} For example, in the context of autonomous vehicles (or other embodied AI systems navigating in a human environment), the safety specification may contain rules about lawful and appropriate driving behaviour as well as specifications to avoid causing physical harm to others; for software systems, the safety specification may seek to capture both its intended functionality and a lack of cyber vulnerabilities; or, in the context of automated medical research, the safety specification could bound the pathological potency of the identified molecules. Section~\ref{sec:examples} provides a number of concrete examples of GS AI, including their safety specifications. 

\begin{figure*}
    \centering
    \includegraphics[width=1\linewidth]{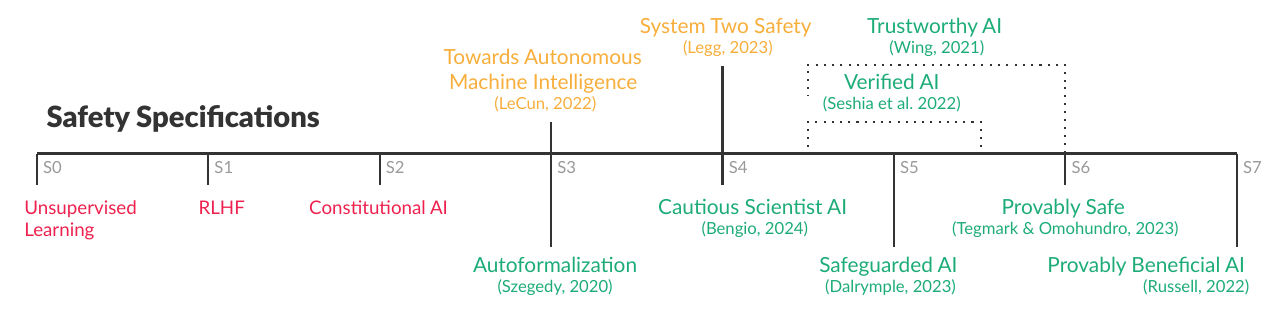}
    \caption{Different approaches for creating safety specifications can be projected onto a spectrum according to how much safety they would grant if successfully implemented. Approaches listed in green fall into the GS AI family. Approaches in yellow may qualify (depending on underspecified aspects of the approach) or come close to qualifying as GS AI. In our view, approaches in red fail to provide high-assurance quantitative safety guarantees, and thus do not qualify as GS AI.}
    \label{fig:specification_spectrum}
\end{figure*}

There are many possible strategies for creating safety specifications. These strategies can roughly be placed on a spectrum, depending on how much safety it would grant if successfully implemented. One way to do this is as follows (see also Figure~\ref{fig:specification_spectrum}):

\begin{itemize}
    \item Level 0: No safety specification is used.
    \item Level 1: The safety of the system is evaluated by a pool of human judges based on their high-level intuitions and preferences.
    \item Level 2: The system uses a safety specification that is expressed in natural language but interpreted by a black-box AI system.
    \item Level 3: The system uses hand-written safety specifications for limited safety properties that are relatively tractable to express in a formal language.
    \item Level 4: The system uses a specification that is written in (probabilistic) logic at the top level, but which makes use of (uninterpreted) neural components to represent learned bindings of certain human concepts to real physical states.
    \item Level 5: The system uses compositional specifications that are made up of parts that are all human audited, but synthesised by AI.
    \item Level 6: The system uses hand-written safety specifications for comprehensive safety properties that require substantial effort to express formally. 
    \item Level 7: The safety specification completely encodes all things that humans might want, in all contexts.
\end{itemize}

Authoring safety specifications is routine in automated controller design \citep{tabuada2009verification}, and can frequently be achieved for both traditional software systems and domain-specific AI. Examples include probabilistic guarantees of collision avoidance in robotics, guarantees that some control policy interacting with a dynamical system will stay in a safe region (used e.g.~in robotics, drones, and controllers for many other physical systems), verification of security properties of cyberphysical systems, or of medical device software. However, creating safety specifications becomes increasingly difficult as AI systems operate in increasingly general and open-ended settings, and without well-defined scopes or semantics. Suppose, for example, that we want to ensure that a chatbot never gives advice that is \enquote{harmful}. How should this specification be formalised? A robust formalisation of this specification would require a very detailed world model, since whether or not a given piece of advice will turn out to be harmful may depend on diverse facts about the real world in complicated ways. Alternatively, we could instead require the AI to never give advice that it \enquote{believes} to be harmful. However, verifying this specification requires a reliable way of extracting \enquote{beliefs} from an AI system, which is difficult for black box models. Moreover, \enquote{harm} is a vague predicate, in the sense that there are edge-cases where it is controversial whether a given person is harmed or not. Similar issues occur if we want to ensure that an AI system never lies, or that it always follows instructions from humans, etc. 

In many cases, it is possible to find \emph{proxies} for complex predicates (such as \enquote{harm}) which are easier to define and measure. 
However, while  such a proxy may robustly correlate with our intuitive judgements of harm in normal situations, they may still reliably come apart if those proxies are used as an optimisation target. This phenomenon is known as \emph{Goodhart's law}, which is an informal principle sometimes stated as \enquote{when a measure becomes a target, it ceases to be a good measure}. Goodhart's law was first introduced by \citet{goodhart_problems_1975}, and has since been studied more formally in works such as \citet{manheim_categorizing_2019, hennessy_goodharts_2023, subset_features, skalse_defining_2022, karwowski2023goodharts}. This means that a safety specification may need to be highly accurate in order to remain robustly reliable, especially when applied to an AI system with strong optimisation or planning capabilities.

One strategy is to attempt to \emph{learn} safety specifications from data. This is explored by the field of \emph{reward learning}, where the specification is assumed to have the form of a reward function. If we assume that a human's preferences can be captured by a reward function, and if we can learn a representation of this reward function from some data source, then we may be able to prove that a given AI system will respect the preferences which are embodied by that reward function. However, reward learning also faces serious difficulties. In particular, most data sources are insufficient for identifying the underlying reward function uniquely, even in the limit of infinite data, and this irreducible ambiguity may be problematic \citep{ng2000, dvijotham2010, cao2021, kim2021, skalse2022invariance, schlaginhaufen2023identifiability}. Moreover, many reward learning algorithms are highly sensitive to the modelling assumptions they make about their data source \citep{armstrong2017, choicesetmisspecification, viano2021robust, skalse2023misspecification, skalse2024quantifying}. Finally, the learnt reward model is itself typically not interpretable, which is a serious issue \citep[see e.g.\ ][]{michaud2020understanding,jenner2022preprocessing}. This means that much work is required before reward learning can be a reliable source of specifications. For an overview, see \citet{casper2023open}. Also note that there are approaches to learning safety specifications which go beyond the umbrella of reward learning, see e.g.\ \citet{bengio2024cautious,seshia-pieee15,vazquez-neurips18,vazquez-cav20,neider2018learning}.

Another approach is to attempt to create \enquote{conservative} safety predicates which aim to be \emph{sufficient} (but not \emph{necessary}) for safety. For example, we may require that the AI has no incentive to influence any part of the external world, or to find out any information about the external world \citep[e.g.\ ][]{armstrong2012thinking, armstrong2018good, armstrong2018indifference, everitt2021agent, vanmerwijk2022complete}. Alternatively, we may attempt to create safety predicates which make an AI system more safe, even if they do not \emph{ensure} safe behaviour \citep[as e.g.\ ][etc]{corrigibility,safelyinterruptibleagents, hadfieldmenell2017offswitch}. These approaches come with their own challenges, see e.g.\ \citet{bostrom2014superintelligence} and the above cited works. Moreover, while certainly challenging, it may still be feasible to simply specify the safety predicates manually. This is (partially) attempted in works such as e.g.\ \citet{harm2022, beckers2022quantifying}.

It is also important to note that the task of formalising specifications can be easier with a \emph{system-level} approach. As advocated by~\citet{seshia-cacm22a}, even when individual AI components perform tasks that are hard to formalise, it can still be the case that the relevant notion of safety can be formalised at the full system level. For example, one can formalise system-level safety for autonomous vehicles even when it is not possible to formalise correctness for object detection and classification components in the autonomy stack~\citep[e.g.,][]{dreossi-jar19}. In some cases, this can make it feasible to define specifications directly.

A complementary approach is to create a hierarchy of safety specifications for {\it provably compliant systems} (PCSs) all the way down to the physical level, as described by \citet{tegmark2023provably}.
Such a hardware-grounded ecosystem of modules is valuable because some of today's formally verified systems have been hacked by compromising physical properties of the underlying hardware \cite{mutlu2019rowhammer}.
To prevent this, the PCS-approach rests on a foundation of relatively simple provably compliant systems, from sensors, actuators and memory devices, to microprocessors with physics-based
proofs that they meet their specification --- including a protocol guaranteed to reveal failure or tampering. Such basic PCS modules are then combined into an ecosystem of increasingly complex PCS’s whose verification recursively verifies their components.

Although specification can be challenging for highly competent and embodied AI systems, there are many low-hanging fruit tasks which do not require complex specification (or world modeling or verification) and which can provide great near-term value and incentives for further development. Examples include machines provably impossible to login to without correct credentials, DNA synthesizers that provably cannot synthesize certain pathogens, and AI hardware that is provably 
geofenced, time-limited (``mortal”) or equipped with a remote-operated throttle or kill-switch. Provably compliant sensors can be spefified to ensure ``zeroization”, in which tampering with PCH is guaranteed to cause detection and erasure of private keys.
For more details, see \citet{tegmark2023provably}.

Another strategy for creating specifications is provided by \emph{Cooperative Inverse Reinforcement Learning} (CIRL), as described by \citet{hadfieldmenell2024cooperative}. CIRL formulates the interaction between an AI and a human as a two-player Markov game (with one player being the AI, and one player being the human). Both players have the same reward function, but only the human knows what the reward function is. This means that the two players must cooperate to obtain a high reward, and that the AI system must listen to feedback from the human. Optimal solutions to the CIRL game produce behaviours such as active teaching, active learning, and communicative actions. The idea is that this problem formulation will encourage the AI to be \emph{corrigible} \citep[in the sense of][]{corrigibility}, instead of following some goal dogmatically. An AI could be verified to adhere to various safety specifications within the CIRL game. These specifications could take several forms, but a natural choice would be to require that the AI is \emph{provably beneficial} to the human (or some variation thereof). Note that such specifications may require the world model to also make modelling assumptions about the behaviour of the human (and in particular about how the behaviour of the human relates to its preferences). Also note that many of the challenges to reward learning also apply to CIRL.

While these challenges are serious, it is important to note that most approaches to AI safety require formal safety specifications (or at least formalisations of what an AI system should be optimised for). These difficulties are thus not unique to the GS research agenda. It is also important to note that further development in AI capabilities will tend to make it easier to create good safety specifications. For example, AI systems could be used to suggest new specifications, to critique proposed specifications, or to generate examples of cases where two candidate specifications differ. Progress in AI could thus also accelerate the creation of good safety specifications.

\subsection{The Verifier}\label{sec:verifier}

In this section we will discuss the verifier, the difficulties with creating such a verifier, and some potential strategies for overcoming these difficulties.

\begin{figure*}
    \centering
    \includegraphics[width=1\linewidth]{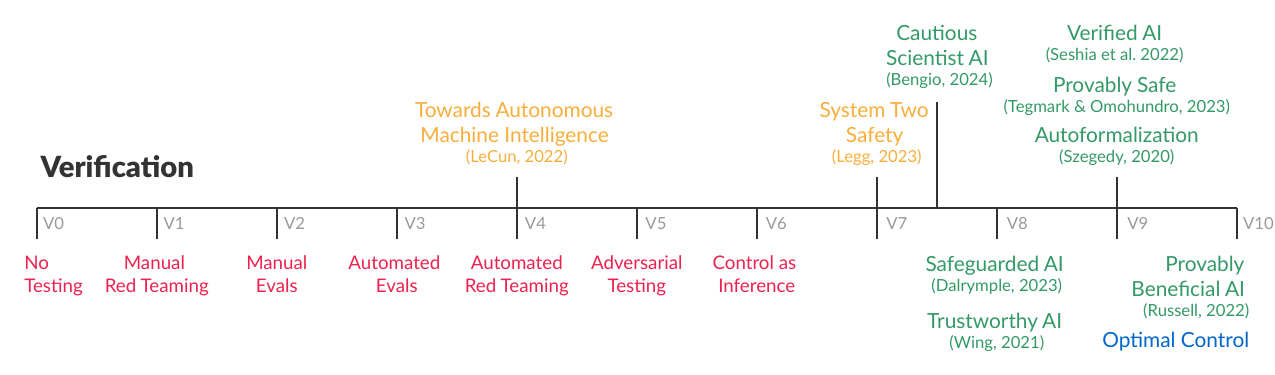}
    \caption{Different approaches to verification can be projected onto a spectrum according to how much safety they would grant if successfully implemented. Approaches listed in green fall into the GS AI family. Approaches in yellow may qualify (depending on underspecified aspects of the approach) or come close to qualifying as GS AI. In our view, approaches in red fail to provide high-assurance quantitative safety guarantees, and thus do not qualify as GS AI.}
    \label{fig:verification_spectrum}
\end{figure*}

The verifier may produce different kinds of formal guarantees, depending on what is feasible in a given context. We can thus place different kinds of verifiers on a spectrum, based on the strength of their corresponding guarantees (see also Figure~\ref{fig:verification_spectrum}):

\begin{itemize}
    \item Level 0: No quantitative guarantee is produced.
    \item Level 1: You use ad-hoc empirical testing of the AI system, which provide a heuristic-based assurance.
    \item Level 2: You use a standardised set of tests, which thus provides safety assurances that are auditable and comparable across systems.
    \item Level 3: You use a property-based test which includes domain randomisation and current state-of-the-art automated evaluations. In other words, you have some template for what the test looks like, but randomly populate the template to get more coverage.
    \item Level 4: You use black-box fuzzing, wherein an automated tool gives test vectors as input to the system and, depending on the response, generates different test vectors to fool the system. In other words, you use a form of automated red-teaming.
    \item Level 5: You use white-box fuzzing, which is akin to Level 4, except that your tool not only looks at the input-output behaviour of the AI system, but also considers its internal states and tries to make certain internal structures flip to get more coverage.
    \item Level 6: You use probabilistic inference with asymptotic convergence. This is akin to Level 5, but with the additional guarantee that if the evaluation system would run forever, then it would eventually literally cover every possible input to the system.
    \item Level 7: You combine asymptotic coverage with white-box fuzzing. This might include adversarial gradient optimisation, whereby you first cover areas where you most expect safety concerns to spring up, and where the system in the limit of infinite time would cover every possible system input.
    \item Level 8: Akin to Level 7, but with the additional requirement that you have some non-asymptotic convergence bounds, due to some formula for how much of the total state space your system covers at a given time. You can thus run it for a finite amount of time, and know how much is left that you have not covered
    \item Level 9: You have a sound bound on the probability of failure, meaning that the true probability of something happening is less than or equal to the value established by your verifier. This includes the formal proof case where the verifier is able to establish that the probability of failure is 0 (relative to the world model).
    \item Level 10: Akin to level 9, but with the additional requirement that the proof is concise enough that humans can read, understand, and check it. \footnote{
    For the formal proof case, level 9 suffices: Just as it is much harder to find a needle in a haystack than to verify that it is a needle, it is much harder to discover an algorithm with a compliance proof than to verify it --- which can be done with just a few hundred lines of human-written code \cite{Metamath}. This means that humans need not understand the system or proof, merely the the simple verifier code.} 
\end{itemize}

To obtain strong guarantees, we need a verifier at Level 8, 9, or 10. This, of course, raises the question of whether such guarantees can feasibly be obtained for AI systems. Even for relatively simple properties of software, obtaining such guarantees currently imposes a very high burden, as it requires highly specialised cognitive labour. However, it may be possible to train AI systems without ASL-4 capabilities to automate much of this labour at a near-expert level of sophistication without raising significant safety concerns. For example, consider an AI agent that
 only interacts with a formal theorem-prover and is only allowed to grow a library of formal facts. In this case the agent will never be exposed to any actions that would meaningfully influence the real world, and it would never learn to communicate with humans outside of formalising and reasoning about existing statements or developing related theories. Note that the scope of training for such an agent would be very limited, i.e., it would only be exposed to natural language mathematics and computer science. It would therefore not have any significant knowledge of the outside world beyond these restricted domains. This restricts the AI’s potential for power-seeking behaviour.

Formal verification has a long and rich history reviewed in {\frenchspacing e.g.} \citet{leino2023program,seligman2023formal}. 
Here we outline a potential approach for designing and training an expert-level reasoning system by letting it process all of the mathematical and computer science literature automatically \citep{szegedy2020promising}. We first index all of the existing informal and formal literature with an ASL-3 system and create a neural retrieval system. Then we use similarly capable language models to extract statements and definitions from the text and formalise them. Early examples of such models already exist, but they must be improved before they are of sufficient quality to be used in the initial phase of a bootstrap loop. The informal to formal translation data for this phase can be generated using cycle-consistency training \citep{lample2017unsupervised} together with the small but increasing amounts of human-verified formalization of informal statements.  Similarly, a proof-generation agent can be trained on the relatively large corpus of existing formal theorems. Additional capability can be achieved via a reinforcement learning loop that trains the translation model and the neural prover in lockstep. In this step, both models can be assumed to be retrieval-augmented large language models, while the translation takes in informal (natural language) specifications and outputs formal specification. Then the prover attempts to prove them, potentially utilising the natural language proof sketches. Whenever a statement is proven to be correct (that is, verified by the formal prover), and it is useful for proving other statements like its informal counterpart, it can be considered to be a correct transcription and used for training the translation model. In addition, the prover can be trained on the successful proof traces. While there is ongoing research work on the overall feedback loop, and many challenges remain, there are multiple existing proof-of-concept solutions for various components of this reinforcement learning system \citep{szegedy2020promising}, which suggests that a full solution may be realizable in the coming years. 

A more extensive discussion on the challenges for extending formal methods to handle the unique characteristics of AI systems is provided by \citet{seshia-cacm22a}, who presents 15 principles for providing provable guarantees of safety for AI systems. We summarise some of the key ideas pertaining to the verifier here and point the reader to \cite{seshia-cacm22a} for details. First of all, one important piece is devising suitable {\em abstractions of AI components} such as neural networks. \citet{seshia-cacm22a} advocate for designing abstractions of AI components that are easier to formally analyse and which can be generated algorithmically. In this context, abstractions that are modular and more interpretable have also been found to be easier to analyse by formal verification tools. Another crucial principle is {\em compositional reasoning}; the idea is to construct a proof of safety of the overall AI system by decomposing the top-level proof obligation into sub-obligations on individual components, and then applying formal verification to the sub-obligations. One challenge here is that not all AI components have formal specifications -- in other words, we need to do compositional verification without compositional specifications! \citet{seshia-cacm22a}  tackle this by proposing techniques to automatically infer a decomposition of the system into components along with {\em interface contracts} between them. Such a decomposition could be spatial and/or temporal and has been shown to aid in scalable analysis~\cite{dreossi-jar19,yalcinkaya-rv23}. Finally, it is crucial to integrate formal methods into the design process for AI components, for example, the training of deep learning components, and also to connect design-time formal methods with run-time assurance.
Ideas from verified inductive synthesis of programs can be useful in training AI components with provable guarantees \citep[e.g.\ ][]{dreossi-ijcai18, abate2023}.

A complementary approach is to replace neural networks by traditional formally verifiable software in deployed systems or key modules thereof \cite{tegmark2023provably}. The algorithms to be coded up and verified can be auto-generated either by direct AI-powered program synthesis or by using automated mechanistic interpretability tools to distill machine-learned algorithms into verifiable code \citep{michaud2024opening}. Both approaches can be accelerated by creation of large benchmarks for training/testing.

\section{Examples of GS AI Solutions
}\label{sec:examples}

Here we describe some concrete examples of practical problems and how they may be solved with a GS AI approach, which combine the different components described in the previous sections and fall along the different axes. The examples range from current- or near-term tractable and economically viable applications, to ones that have not yet been realised due to their more ambitious scope. 

\begin{mdframed}
\begin{example}[Code Translation]\ \\
\textbf{Problem}: Translate code in one programming language (e.g.\ C) to another (e.g.\ Rust) using AI systems (e.g.\ language models) with provable guarantees.
While some simple instances of this general problem have been solved (e.g., translating side effect-free functional programs from a source language to a domain-specific target language~\cite{bhatia2024verified}), much remains to be done to solve the general code translation problem.

\textbf{Specification}: Functional correctness (the source program is equivalent to target program with respect to the relevant program state), and the generated target code does not have specified classes of security vulnerabilities.
Defining functional correctness is easy in some cases (e.g.,~\cite{bhatia2024verified}) but more complicated in general since the source and target program structure can be very different and specifications can be partial.

\textbf{World Model}: Constraints on the inputs to the program (pre-conditions), and threat models for security (what the attacker can/cannot do).

\textbf{AI System}:  Large language models for code generation (e.g.\ models such as GPT-4, Claude, etc., or others expressly designed or fine-tuned for code generation).   
\end{example}
\end{mdframed}

\begin{mdframed}
\begin{example}[Autonomous Vehicle (AV) Safety]\ \\
\textbf{Problem}: Design an AI-enabled autonomous driving system (ADS) that satisfies a formal specification of safety and functionality in specified scenarios, and also maintains a (smaller) core safety specification when outside those scenarios 

\textbf{Specification}: A \enquote{rulebook}~\cite{censi-icra19}, a set of rules that have an associated pre-order, where each rule can be a specification written in a formalism that suits its purpose. Rules are mathematical objects that map behaviours (or sets of behaviours) to a value domain; they can be objective functions (cost/reward functions), temporal logic properties, probabilistic bound properties, automata, etc. The overall specification would also include a \emph{more restricted} specification that would be maintained by the backup system (of which there could be multiple levels).

\textbf{World Model}: A set of scenarios in which the ADS operates that includes formal models of (i) the other components of the AV, (ii) agents and objects in the scenario, including their attributes and behaviour models, (iii) other attributes of the environment, such as spatial constraints of the physical world around the AV, time of day, weather conditions, etc., (iv) models of humans and their preferences and objectives. We expect these models will be stochastic in nature and some components of the world model will be learned from data/interactions while others will be encoded by domain experts or inferred/generated statically. Probabilistic programming languages \citep[e.g.\ Scenic,][]{scenic-mlj22} offer one mechanism for world modelling for this application and have had some success in practice.  As in the case of the specification, there would be world models also specified for the backup system(s) which constitute weaker assumptions about the operating environment.

\textbf{AI System}: ADS integrates many types of AI/ML components with traditional hardware and software components. For example, deep neural networks are used in various steps of the autonomy pipeline including for perception, prediction, planning, and control. LLMs are being considered for use in interfaces to humans, to elicit preferences, explain the behaviour of AVs to occupants and others, etc. An ADS should also include runtime monitors that check whether the assumptions made in the world model (about the AV’s “operational design domain”) are maintained during operation, and, if not, to predictively take backup action. 
\end{example}
\end{mdframed}

\begin{mdframed}
\begin{example}[Household Robots]\ \\
\textbf{Problem}: Provide assistance to humans in need of support at home.

\textbf{Specification}: Carrying out diverse household tasks. Tasks include cleaning rooms, organising items, bringing items to humans, and navigating safely around people, ensuring no harm comes to humans. Adaptation to a human's special needs and environment: the robots should adapt to varying human needs and environments while grounding the original skills to the new conditions.

\textbf{World model}: Home diversity (various layouts of homes and diverse tasks for home management) and robot constraints (constraints governing the behaviour of the robots to ensure safety and efficiency).

\textbf{AI System}: VLMs equipped with multi-task learning capabilities and continuous learning mechanisms.
\end{example}
\end{mdframed}

\begin{mdframed}
\begin{example}[Medical Diagnosis Agent]\ \\
\textbf{Problem}: Facilitate disease diagnosis and provide medical advice.

\textbf{Specification}: Precisely diagnose diseases based on the presenting symptoms of the patient, provide insights into the underlying causes of the disease, and offer precise and effective treatment recommendations.

\textbf{World Model}: A model of the human body (describing the complexities and nuances of human physiology and pathology), and a model of environmental factors (for consideration of external factors contributing to disease onset or progression).

\textbf{AI System}: A multi-media input system capable of parsing diagnostic notes, interpreting medical imaging data, and delivering accurate diagnoses and treatment plans.
\end{example}
\end{mdframed}

\begin{mdframed}
\begin{example}[Compute Centre Management]\ \\
\textbf{Problem}: Ensuring that data centres run multi-tenant applications efficiently and correctly while preventing unauthorised applications from using compute.

\textbf{Specification}: Efficient allocation according to business agreements, legal compliance with e.g.\ compute caps, and security (protection against and detection of attempted breaches).

\textbf{World Model}: The shared computing ecosystem with tenant isolation, resource allocation, and cryptographically secured hardware.

\textbf{AI System}: An AI model trained for optimal allocation within economic, legal, and security constraints.
\end{example}
\end{mdframed}

\begin{mdframed}
\begin{example}[Formally Verified Re-Implementation of Bug-Free Software and Hardware Systems for Cyber Defence.]\ \\
\textbf{Problem}: The currently existing soft- and hardware infrastructure contains many bugs, thereby creating a large attack surface area for cyberattacks. A drastic reduction of this surface area and thus improved cybersecurity can be achieved by replacing legacy code with software that has been formally verified to be bug-free and trustworthy.

\textbf{Specification}: The intended function of the legacy code, and a lack of bugs or vulnerabilities.

\textbf{World Model}: Autoformalisation taking the legacy code and inferring from it its intended function.

\textbf{AI System}: An AI \enquote{Software Engineer} agent which re-implements existing software through autoformalisation and interactive specification crafting with stakeholders, and synthesises a bug-free version. 
\end{example}
\end{mdframed}

\begin{mdframed}
\begin{example}[Safe Nucleic Acid Sequence Screening \& Synthesis]\ \\
\textbf{Problem}: The screening, identification and synthesis of nucleic acid, protein, RNA, or DNA structures has potent applications for the health and flourishing of humans, however also risks applications that could be harmful for individuals or civilisation at large. 

\textbf{Specification}: The rejection for synthesis of sequences that could be used in the production of pathogens, toxic chemicals, or other substances potent to negatively affect human health, and the identification and acceptance of the synthesis of sequences with potential for applications that are conducive to human health and flourishing.

\textbf{World Model}: The relationship between molecular structures and pathology.

\textbf{AI System}: A \enquote{bioengineering agent} able to verify, for a given sequence, whether its risk of harmful applications remains below a conservatively specified bound.
\end{example}
\end{mdframed}

\begin{mdframed}
\begin{example}[Stabilisation of the Climate System]\ \\
\textbf{Problem}: The emission of greenhouse gases threatens to throw out of balance the climate system, leading to increasingly averse climate conditions and threatening to destroy the health and livability of the ecosystem.  

\textbf{Specification}: Ensure the stability of the climate system, and the health of the ecosystem under said climate.

\textbf{World model}: A model of complex climate dynamics under various intervention regimes.

\textbf{AI Agent}: A \enquote{climate agent} that generates geoengineering solutions which verifiably stabilise the climate system without compromising the health of the ecosystem.
\end{example}
\end{mdframed}

\section{Discussion}\label{sec:discussion}

We have argued that safety-critical AI systems need to be equipped with quantitative safety guarantees, and that empirical evaluations alone are insufficient for producing such guarantees with an adequate level of assurance. We have also introduced GS AI as a potential avenue for feasibly obtaining such safety guarantees, discussed the main challenges with this research agenda, and some ways in which those challenges may be overcome. In this section, we will further discuss some potential benefits to the GS AI research agenda, and some important considerations.

An important benefit to GS AI is that it makes democratic oversight easier, because concrete safety specifications can be audited and discussed by outside observers and regulators. AI technology will have an immense (and potentially unprecedented) impact on most, if not all, areas of life. This makes it crucial to enable as many people as possible to have a say in how this technology is deployed and used. However, such democratic input and oversight is not possible if important AI systems are specified, trained, and evaluated using procedures that are opaque to (or kept secret from) the wider public. In other safety-critical industries, probabilistic safety assessment means that the developer must specify all the assumptions or premises needed to deduce that their system meets the required societal risk thresholds. These assumptions can then be challenged by the regulator (and society at large) if they are not deemed socially acceptable. GS AI enables the same kind of oversight for AI systems.

A further socio-technical benefit of GS AI is its potential for facilitating multi-stakeholder coordination. This is because GS AI is able to produce proof certificates verifying that a given solution conforms to auditable-to-all-parties specifications, in a way that requires minimal trust among said parties. It is well known among scholars and practitioners in International Relations that such a low-trust verification ability is a critical ingredient to multi-stakeholder coordination. For example, the successful ratification of several international nuclear treaties has, to a significant extent, been made possible because the respective parties are able to verify that each the others’ nuclear activities are not being retargeted to military purposes. 

A third benefit of the GS research agenda is that it may produce AI safety solutions whose costs are amortised over time. Any potential safety measure (in AI or elsewhere) faces the issue that if said safety measure is costly to implement, then there is an incentive to disregard it. This reduces the value of approaches to AI safety that impose a high \enquote{safety tax}. For example, a comprehensive suite of rigorous empirical evaluations may be expensive and time-consuming to carry out, and this cost would (presumably) have to be paid again for each new AI system that is created. This would in turn create an incentive to save resources by cutting corners. By contrast, some approaches to GS AI may allow for most of these costs to be amortised, which may substantially reduce the incentive to disregard the corresponding safety measures. Once satisfactory safety specifications have been identified and scalable methods for formal verification have been developed, new AI systems could likely be verified against these specifications at a much lower marginal cost. Such solutions would also be more scalable in view of the impressive speed of AI progress that otherwise threatens to outpace progress in safety.

The range of applications for GS AI solutions, and with that also the scope of potential benefits, is large. A particularly significant type of application, which seems thus worth highlighting in the current context, is the securing of critical infrastructure such as water supply, power grids, nuclear plants, dams, information and communication technology, etc. Cybervulnerability of critical infrastructure may be one of the biggest threat vectors for catastrophic risks in the current world. For example, the US Cyber Security and Infrastructure Security Agency (CISA) states that the \enquote{incapacitation or destruction [of critical infrastructure] would have a debilitating effect on security, national economic security, national public health or safety, or a combination thereof}.\footnote{Retrieved from \url{https://www.cisa.gov/topics/critical-infrastructure-security-and-resilience/critical-infrastructure-sectors} on May 10, 2024.} 
GS AI approaches have the potential to play a critical role in securing such critical infrastructure, thereby drastically bolstering civilisational resilience to both 
accidents and adversarial threats.

We also want to emphasise that there is a spectrum of approaches for safety assessments, ranging from easy ones that provide weaker safety assurances (such as evaluations and red teaming) to more expensive procedures that provide stronger safety guarantees (such as the approaches within the GS AI spectrum). Given the uncertainty about when each of the ASL safety levels will be crossed, we need an \enquote{anytime} portfolio approach of R\&D efforts spanning this spectrum. This will allow us to maximise the expected effectiveness of the feasible safety techniques at each stage. This would involve investing in cheaper techniques, such as empirical evaluations, but also more ambitious approaches, such as those presented by the GS AI framework. Furthermore, quick partial successes in GS AI are both plausible and useful. For example, provably compliant cybersecurity, geofencing, remote kill switches, verified sensors and actuators, etc., may all be feasible even if the most ambitious proposals within the GS AI agenda are not, and would still provide benefits in their own right. Beyond the need for the best available protections against risks at any point in time, it is also advisable to seek \enquote{defence in depth} --  a term often used in safety engineering and cybersecurity, advocating for the idea of deploying multiple layers of protection in order to defend against complex threats. 

In this spirit, an important further aspect of a GS AI in practice is its deployment infrastructure. In particular, it might be desirable to monitor the AI system at runtime, and check for signs that the world model is inaccurate. If such signs are detected then the AI system could be disabled, to ensure that there is no unaccounted for risk of unsafe behaviour. It is worth noting that in cases where a system is safety critical, it may not be possible or desirable to simply shut it down. For example, it would not be safe to disable the controller of a self-driving car while the car is in use. A GS AI system should therefore have a trustworthy backup system, appropriate for its domain of use, that can safely transition the system to a safe state if the main AI controller is disabled. Enabling such rapid automatic response to observations that invalidate the predictions of the world model, the deployment infrastructure can therefore grant additional safety.
Also note that GS AI systems themselves may be useful for creating bug-free software and tamper-proof hardware that could be used in backup systems. 

Via the deployment infrastructure, we can modulate the time horizon at which the verified AI output is deployed. For example, at one level, the AI system may be deployed to operate autonomously with an infinite time horizon. To increase the reliability of safety guarantees, one may instead choose to deploy an AI system for a finite time horizon, such that, if at the end of that time horizon the AI system is not successfully verified for redeployment, operations shut down or transition to a safe mode. The deployment systems can therefore also allow us to shorten the time horizon of outputs that need to be verified, thereby reducing the complexity of verification and increasing its tractability.

Moreover, while we have argued for the need for verifiable quantitative safety guarantees, it is important to note that GS AI may not be the only route to achieving such guarantees. An alternative approach might be to extract interpretable policies from black-box algorithms via automated mechanistic interpretability and directly proving safety guarantees about these policies. This approach differs from GS AI in that it does not make use of a world model that is separate from the policy; instead, it requires that the policy can itself be made interpretable. This strategy may be easier if it is intractably difficult to create a sufficiently good world model or adequate methods for doing formal verification relative to that world model. However, it may also be more difficult, especially because the policy may be more complex than the world model. For example, the rules of chess are less complex than a policy which is good at chess, and it is much easier to specify the axioms of Euclidean geometry than it is to specify a computer program that is good at proving theorems about Euclidean geometry. In a similar way, it may be much easier to create an interpretable world model than to create a performant interpretable policy. However, it is ultimately an empirical question whether it is easier to create interpretable world models or interpretable policies in a given domain of operation.

\section{Relevant Existing Work}\label{sec:existing_work}

In this section, we briefly provide an overview on some existing work that GS AI approaches build on, or which are otherwise relevant to or related to the GS AI agenda. Note that we do not mean to claim novelty with respect to the three-component architecture - for example, arguably the first system with separate components for world-modelling, specification, and verification was the (GPS), published by \citep{gps1961}. Instead, this position paper seeks to identify the convergence between several existing approaches, provide a taxonomy, and outline a high-level research agenda towards achieving high-assurance quantitative safety guarantees.

\textbf{Computational Learning Theory.} The field of computational learning theory (CLT) is concerned with the mathematical analysis of learning algorithms and learning problems. These investigations have produced various formal guarantees for large classes of learning algorithms, primarily in the form of generalisation guarantees and regret bounds. These bounds typically show that a given learning algorithm under some given circumstances is guaranteed to attain a given level of performance with at least some given probability, and may also show how these probabilities scale in terms of the amount of training data, etc \citep[see e.g.\ ][]{introtoCLT}. Like the GS approach to AI safety, CLT is also concerned with deriving formal guarantees for AI systems. However, CLT is typically concerned with guarantees concerning narrow performance metrics, whereas we are additionally concerned with specifications that would provide strong safety assurances for advanced systems operating in open-ended environments.

\textbf{Formal Verification of AI Systems.} There is also a large literature on formal verification of AI-enabled systems. This builds on more classic work on formal verification of control policies (e.g., \citep{tabuada2009verification}) and classical expert systems \citep{o1993expert}, to develop methods for formal verification of neural networks and other machine learning systems.
This literature has proposed a range of algorithms which can be used to ensure that a given neural network (potentially belonging to some restricted class) satisfies a given specification (typically specified over its interfaces), or that a system containing 
neural components satisfies a safety specification. It would not be possible for us to provide a comprehensive overview of these algorithms here, but for an introduction, see e.g.\ \citet{seshia-cacm22a, albarghouthi2021introduction}.

\textbf{Program Synthesis.} Another area of computer science that is relevant to GS AI is the literature on \emph{correct-by-construction} program synthesis \cite{alur-fmcad13,ABCCDKKP20}. This is an approach to software development that aims to produce programs that are guaranteed to be correct with respect to their specifications from the outset, rather than relying on traditional testing and debugging methods to identify and fix errors after the fact. The core idea behind correct-by-construction synthesis is to use formal methods to systematically construct programs that satisfy a given set of formal specifications or correctness properties. This is typically achieved by encoding the specifications as logical formulas or constraints, and then using automated solvers or synthesis algorithms to derive programs that provably satisfy those constraints. Many contemporary program synthesis methods involve learning algorithms, e.g., using neural network heuristics, guided by oracles such as formal verifiers or LLMs to accerelate the learning process \citep{ellis2023dreamcoder,grand2023lilo}. For an overview, see e.g.\ \citet{PGL-010,jha-acta17,edwards2023general}.

\textbf{Model-Based Reasoning and Planning.} Related to program synthesis are the fields of automated theorem proving \citep{loveland2016automated} and automated planning \citep{ghallab2004automated}. Both areas study methods for efficient automated reasoning over formal models, with the former focused on reasoning over mathematical theories and program logics, and with the latter, by analogy \citep{fikes1971strips}, focused on reasoning over theories of physical or virtual worlds. Automated theorem proving is important for GS AI due to the role it plays in the verifier component, while automated planning can be used to deduce plans (rather than programs) which are correct-by-construction. This means that automated planners can serve as self-verifying AI systems by generating plans that both reach a goal and satisfy a formal safety specification. A limitation of classical automated planners is that they operate over deterministic world models which only admit Boolean variables. However, planning algorithms have also been extended to work with (partially observable) stochastic world models \citep{bonet2009solving}, with correctness maintained in the form of asymptotic convergence guarantees \citep{bonet2003labeled} or guaranteed bounds \citep{McMahan2005}.

\textbf{Probabilistic Programming.} To specify or learn world models which are interpretable and expressive enough for GS AI, an appropriately rich formal representation is necessary. Probabilistic programming provides one such formal representation, allowing world models to be specified as (possibly Turing-complete) programs that integrate probabilistic uncertainty, symbolic code, and learned differentiable components \citep{blog,goodman2012church,bingham2019pyro,cusumano2019gen,fremont-pldi19}. Recent advances in probabilistic programming have also led to more general and scalable methods for Bayesian inference \citep{etalumis,lew2023smcp3}, probabilistic program synthesis \citep{saad2023sequential}, and automatic differentiation of expected values \citep{lew2023adev}, which may serve as key enablers for learning and refining world models from data.

\textbf{Safety Engineering.} GS AI is more broadly related to the field of safety engineering, an area of engineering science that focuses on identifying, evaluating, and mitigating potential hazards and risks associated with various systems and processes. For an overview of the techniques used in this field, see e.g.\ \citet{ericson2015hazard,engineering_a_safer_world,dhillon2003engineering}. See also \citet{khlaaf2023toward} for a discussion of safety engineering concepts in the context of AI safety, and how ensuring the safety of AI systems differs from the concept of AI value alignment.

\textbf{Boxing Methods.} A core feature of GS AI is that the AI system is prevented from interacting with the world \enquote{directly}, in the sense that any output must be approved by the verifier.\footnote{Though note that both static and runtime verification are part of the GS AI spectrum.} This may remind some readers of work exploring the idea of \enquote{boxing} a powerful AI system to prevent it from causing harm \citep{armstrong2012thinking, armstrong2018good, armstrong2017low, turchin2021multilevel}. This raises the question how GS AI is similar and/or different from such \enquote{AI boxing}, and whether it suffers from the same challenges \citep[e.g.\ ][]{Pepp2022-PEPMMB}. While there is a similarity between these approaches, a key difference is that, while prior AI boxing proposals tend to seek to limit the AI’s ability to affect the world by limiting its type of interaction channels, the \enquote{gate} implemented by the GS AI architectures is the formal verification of the AI’s output.

\section{Conclusion}\label{sec:conclusion}

In this paper, we have introduced and defined the concept of guaranteed safe (GS) AI. GS AI aims to ensure the safety of AI systems by equipping them with formal, verifiable and auditable safety guarantees. 

GS AI faces serious technical challenges. Creating accurate and interpretable world models, formulating precise safety specifications, and performing formal verification at scale are all difficult problems. However, we have suggested potential strategies and research directions for making progress on these problems. Furthermore, the GS approach may be  necessary given the serious  limitations of other methods such as empirical testing and interpretability. That said, given the complexity and importance of the challenge ahead, we think that these different approaches should be seen as  comprising a portfolio of complementary R\&D efforts, which can be combined in ways that improve on the overall safety of a system. 

We believe the GS agenda is crucial for ensuring robust and reliable safety in advanced AI systems. While empiricism and interpretability are useful tools, they do not provide the strong safety assurances that formal verification can. Although formal verification is challenging, the GS research program offers a promising path toward making it feasible at scale. Value is to be had from narrow or partial solutions all the way to more ambitious applications of GS AI.

Much work remains to fully develop the GS approach. But given its importance for avoiding AI risks, we argue that the GS agenda deserves substantially more attention and resources than it currently receives. With a concerted research effort on the core technical problems, significant progress could be made. We hope this paper provides a useful starting point and motivation for a wider pursuit of the GS program.

\section{Author contributions and Acknowledgements}\label{sec:acknowledgments}
David 'davidad' Dalrymple and Joar Skalse were the lead drafters and authors. Yoshua Bengio, Stuart Russell, Max Tegmark, Sanjit A. Seshia, Steve Omohundro and Christian Szegedy are core authors. Contributing authors include Alessandro Abate, Joseph Y. Halpern, Clark Barrett, Ding Zhao, Tan Zhi-Xuan, Jeannette Wing, Joshua B. Tenenbaum, Nora Ammann, and Ben Goldhaber. Ben Goldhaber and Nora Ammann led the coordination of the project. 

We would further like to thank Anca Dragan, Adam Gleave, Adam Marblestone, Zac Hatfield-Dodds, Heidy Khlaaf, Adam Marblestone, and Vikash Mansinghka for their valuable contributions and feedback on the idea and paper.

Yoshua Bengio, Stuart Russell, Max Tegmark, and David 'davidad' Dalrymple would like to thank the UK AI Safety Institute for hosting the Bletchley Summit, where they obtained common knowledge of their AI safety agendas’ similarities, which was a key causal factor in this paper’s creation.

David 'davidad' Dalrymple would like to thank Eric Drexler for formulating the Open Agency Model~\cite{drexler23}, which previously conceptualised safe AI systems as composed of interacting parts with separation of concerns between specification, problem-solving, and solution-checking.

\bibliography{bibliography}

\begin{thebibliography}{214}
\providecommand{\natexlab}[1]{#1}
\providecommand{\url}[1]{\texttt{#1}}
\expandafter\ifx\csname urlstyle\endcsname\relax
  \providecommand{\doi}[1]{doi: #1}\else
  \providecommand{\doi}{doi: \begingroup \urlstyle{rm}\Url}\fi

\bibitem[Abate et~al.(2020)Abate, Bessa, Cattaruzza, Cordeiro, David, Kesseli, Kroening, and Polgreen]{ABCCDKKP20}
Abate, A., Bessa, I., Cattaruzza, D., Cordeiro, L., David, C., Kesseli, P., Kroening, D., and Polgreen, E.
\newblock Automated formal synthesis of provably safe digital controllers for continuous plants.
\newblock \emph{Acta Informatica}, 57\penalty0 (3):\penalty0 223–244, 2020.

\bibitem[Abate et~al.(2023)Abate, Edwards, Giacobbe, Punchihewa, and Roy]{abate2023}
Abate, A., Edwards, A., Giacobbe, M., Punchihewa, H., and Roy, D.
\newblock Quantitative verification with neural networks.
\newblock Schloss Dagstuhl – Leibniz-Zentrum für Informatik, 2023.
\newblock \doi{10.4230/LIPICS.CONCUR.2023.22}.
\newblock URL \url{https://drops.dagstuhl.de/entities/document/10.4230/LIPIcs.CONCUR.2023.22}.

\bibitem[Alaga \& Schuett(2023)Alaga and Schuett]{alaga2023coordinated}
Alaga, J. and Schuett, J.
\newblock Coordinated pausing: An evaluation-based coordination scheme for frontier ai developers, 2023.

\bibitem[Albarghouthi(2021)]{albarghouthi2021introduction}
Albarghouthi, A.
\newblock Introduction to neural network verification, 2021.

\bibitem[Alpern \& Schneider(1987)Alpern and Schneider]{AlpernSchneider87}
Alpern, B. and Schneider, F.~B.
\newblock Recognizing safety and liveness.
\newblock \emph{Distributed Comput.}, 2\penalty0 (3):\penalty0 117--126, 1987.

\bibitem[Alur et~al.(2013)Alur, Bodik, Juniwal, Martin, Raghothaman, Seshia, Singh, Solar-Lezama, Torlak, and Udupa]{alur-fmcad13}
Alur, R., Bodik, R., Juniwal, G., Martin, M. M.~K., Raghothaman, M., Seshia, S.~A., Singh, R., Solar-Lezama, A., Torlak, E., and Udupa, A.
\newblock Syntax-guided synthesis.
\newblock In \emph{Proceedings of the IEEE International Conference on Formal Methods in Computer-Aided Design (FMCAD)}, pp.\  1--17, October 2013.

\bibitem[Amodei(2023)]{testimony_amodei}
Amodei, D.
\newblock Written testimony of {Dario Amodei} before the {U.S. Senate Committee on the Judiciary, Subcommitee on Privacy, Technology, and the Law}, 2023.
\newblock URL \url{https://www.judiciary.senate.gov/imo/media/doc/2023-07-26_-_testimony_-_amodei.pdf}.

\bibitem[Amodei et~al.(2016)Amodei, Olah, Steinhardt, Christiano, Schulman, and Man{\'e}]{amodei2016concrete}
Amodei, D., Olah, C., Steinhardt, J., Christiano, P., Schulman, J., and Man{\'e}, D.
\newblock Concrete problems in ai safety.
\newblock \emph{arXiv preprint arXiv:1606.06565}, 2016.

\bibitem[Anthropic(2023)]{anthropic_responsible_scaling}
Anthropic.
\newblock Anthropic's responsible scaling policy, 2023.
\newblock URL \url{https://www-cdn.anthropic.com/1adf000c8f675958c2ee23805d91aaade1cd4613/responsible-scaling-policy.pdf}.

\bibitem[Armstrong \& Levinstein(2017)Armstrong and Levinstein]{armstrong2017low}
Armstrong, S. and Levinstein, B.
\newblock Low impact artificial intelligences, 2017.

\bibitem[Armstrong \& Mindermann(2018)Armstrong and Mindermann]{armstrong2017}
Armstrong, S. and Mindermann, S.
\newblock Occam's razor is insufficient to infer the preferences of irrational agents.
\newblock In \emph{Proceedings of the 32nd International Conference on Neural Information Processing Systems}, volume~31, pp.\  5603--5614, Montr\'{e}al, Canada, 2018. Curran Associates, Inc., Red Hook, NY, USA.

\bibitem[Armstrong \& O'Rorke(2018)Armstrong and O'Rorke]{armstrong2018good}
Armstrong, S. and O'Rorke, X.
\newblock Good and safe uses of ai oracles, 2018.

\bibitem[Armstrong \& O'Rourke(2018)Armstrong and O'Rourke]{armstrong2018indifference}
Armstrong, S. and O'Rourke, X.
\newblock 'indifference' methods for managing agent rewards, 2018.

\bibitem[Armstrong et~al.(2012)Armstrong, Sandberg, and Bostrom]{armstrong2012thinking}
Armstrong, S., Sandberg, A., and Bostrom, N.
\newblock Thinking inside the box: Controlling and using an oracle ai.
\newblock \emph{Minds and Machines}, 22:\penalty0 299--324, 2012.

\bibitem[Asimov(1942)]{asimov1942}
Asimov, I.
\newblock Runaround.
\newblock \emph{Astounding Science Fiction}, 1942.

\bibitem[Baier \& Katoen(2008)Baier and Katoen]{Baier2008}
Baier, C. and Katoen, J.-P.
\newblock \emph{Principles of Model Checking}.
\newblock The MIT Press, 2008.
\newblock ISBN 026202649X.

\bibitem[Baydin et~al.(2019{\natexlab{a}})Baydin, Shao, Bhimji, Heinrich, Meadows, Liu, Munk, Naderiparizi, Gram-Hansen, Louppe, Ma, Zhao, Torr, Lee, Cranmer, Prabhat, and Wood]{Baydin2019}
Baydin, A.~G., Shao, L., Bhimji, W., Heinrich, L., Meadows, L., Liu, J., Munk, A., Naderiparizi, S., Gram-Hansen, B., Louppe, G., Ma, M., Zhao, X., Torr, P., Lee, V., Cranmer, K., Prabhat, and Wood, F.
\newblock Etalumis: bringing probabilistic programming to scientific simulators at scale.
\newblock In \emph{Proceedings of the International Conference for High Performance Computing, Networking, Storage and Analysis}, SC '19, New York, NY, USA, 2019{\natexlab{a}}. Association for Computing Machinery.
\newblock ISBN 9781450362290.
\newblock \doi{10.1145/3295500.3356180}.
\newblock URL \url{https://doi.org/10.1145/3295500.3356180}.

\bibitem[Baydin et~al.(2019{\natexlab{b}})Baydin, Shao, Bhimji, Heinrich, Meadows, Liu, Munk, Naderiparizi, Gram-Hansen, Louppe, Ma, Zhao, Torr, Lee, Cranmer, Prabhat, and Wood]{etalumis}
Baydin, A.~G., Shao, L., Bhimji, W., Heinrich, L., Meadows, L., Liu, J., Munk, A., Naderiparizi, S., Gram-Hansen, B., Louppe, G., Ma, M., Zhao, X., Torr, P., Lee, V., Cranmer, K., Prabhat, and Wood, F.
\newblock Etalumis: bringing probabilistic programming to scientific simulators at scale.
\newblock In \emph{Proceedings of the International Conference for High Performance Computing, Networking, Storage and Analysis}, SC '19, New York, NY, USA, 2019{\natexlab{b}}. Association for Computing Machinery.
\newblock ISBN 9781450362290.
\newblock \doi{10.1145/3295500.3356180}.
\newblock URL \url{https://doi.org/10.1145/3295500.3356180}.

\bibitem[Beckers et~al.(2022{\natexlab{a}})Beckers, Chockler, and Halpern]{harm2022}
Beckers, S., Chockler, H., and Halpern, J.
\newblock A causal analysis of harm.
\newblock In Koyejo, S., Mohamed, S., Agarwal, A., Belgrave, D., Cho, K., and Oh, A. (eds.), \emph{Advances in Neural Information Processing Systems}, volume~35, pp.\  2365--2376. Curran Associates, Inc., 2022{\natexlab{a}}.
\newblock URL \url{https://proceedings.neurips.cc/paper_files/paper/2022/file/100c1f131893d3b4b34bb8db49bef79f-Paper-Conference.pdf}.

\bibitem[Beckers et~al.(2022{\natexlab{b}})Beckers, Chockler, and Halpern]{beckers2022quantifying}
Beckers, S., Chockler, H., and Halpern, J.~Y.
\newblock Quantifying harm, 2022{\natexlab{b}}.

\bibitem[Bengio(2023)]{testimony_bengio}
Bengio, Y.
\newblock Written testimony of {Yoshua Bengio} before the {U.S. Senate Committee on the Judiciary, Subcommitee on Privacy, Technology, and the Law}, 2023.
\newblock URL \url{https://www.judiciary.senate.gov/imo/media/doc/2023-07-26_-_testimony_-_bengio.pdf}.

\bibitem[Bengio(2024)]{bengio2024cautious}
Bengio, Y.
\newblock Towards a cautious scientist {AI} with convergent safety bounds, 2024.
\newblock URL \url{https://yoshuabengio.org/2024/02/26/towards-a-cautious-scientist-ai-with-convergent-safety-bounds/}.

\bibitem[BenThacker et~al.(2006)BenThacker, Andereson, Senseny, and Rodriguez]{thacker2006VV}
BenThacker, Andereson, C., Senseny, P., and Rodriguez, E.
\newblock The role of nondeterminism in model verification and validation.
\newblock \emph{International Journal of Materials and Product Technology}, 25\penalty0 (1-3), 2006.

\bibitem[Bhatia et~al.(2024)Bhatia, Qiu, Hasabnis, Seshia, and Cheung]{bhatia2024verified}
Bhatia, S., Qiu, J., Hasabnis, N., Seshia, S.~A., and Cheung, A.
\newblock Verified code transpilation with {LLMs}.
\newblock 2024.

\bibitem[Bingham et~al.(2019)Bingham, Chen, Jankowiak, Obermeyer, Pradhan, Karaletsos, Singh, Szerlip, Horsfall, and Goodman]{bingham2019pyro}
Bingham, E., Chen, J.~P., Jankowiak, M., Obermeyer, F., Pradhan, N., Karaletsos, T., Singh, R., Szerlip, P., Horsfall, P., and Goodman, N.~D.
\newblock Pyro: Deep universal probabilistic programming.
\newblock \emph{Journal of machine learning research}, 20\penalty0 (28):\penalty0 1--6, 2019.

\bibitem[Bloomfield et~al.(2019)Bloomfield, Khlaaf, Conmy, and Fletcher]{khlaaf2019assuringMLauto}
Bloomfield, R., Khlaaf, H., Conmy, P.~R., and Fletcher, G.
\newblock Disruptive innovations and disruptive assurance: Assuring machine learning and autonomy.
\newblock \emph{Computer}, 52/9:\penalty0 82--89, 2019.

\bibitem[Bonet \& Geffner(2003)Bonet and Geffner]{bonet2003labeled}
Bonet, B. and Geffner, H.
\newblock Labeled rtdp: Improving the convergence of real-time dynamic programming.
\newblock In \emph{ICAPS}, volume~3, pp.\  12--21, 2003.

\bibitem[Bonet \& Geffner(2009)Bonet and Geffner]{bonet2009solving}
Bonet, B. and Geffner, H.
\newblock Solving pomdps: Rtdp-bel vs. point-based algorithms.
\newblock In \emph{IJCAI}, pp.\  1641--1646. Pasadena CA, 2009.

\bibitem[Bostrom(2002)]{bostrom2002}
Bostrom, N.
\newblock Existential risks: Analyzing human extinction scenarios and related hazards.
\newblock \emph{Journal of Evolution and technology}, 9, 2002.

\bibitem[Bostrom(2014)]{bostrom2014superintelligence}
Bostrom, N.
\newblock \emph{Superintelligence: Paths, Dangers, Strategies}.
\newblock Oxford University Press, 2014.
\newblock ISBN 9780199678112.
\newblock URL \url{https://books.google.com/books?id=7_H8AwAAQBAJ}.

\bibitem[Boxell et~al.(2020)Boxell, Gentzkow, and Shapiro]{NBERw26669}
Boxell, L., Gentzkow, M., and Shapiro, J.~M.
\newblock Cross-country trends in affective polarization.
\newblock Working Paper 26669, National Bureau of Economic Research, January 2020.
\newblock URL \url{http://www.nber.org/papers/w26669}.

\bibitem[Bricken et~al.(2023)Bricken, Templeton, Batson, Chen, Jermyn, Conerly, Turner, Anil, Denison, Askell, et~al.]{bricken2023towards}
Bricken, T., Templeton, A., Batson, J., Chen, B., Jermyn, A., Conerly, T., Turner, N., Anil, C., Denison, C., Askell, A., et~al.
\newblock Towards monosemanticity: Decomposing language models with dictionary learning.
\newblock \emph{Transformer Circuits Thread}, pp.\ ~2, 2023.

\bibitem[Broeck et~al.(2011)Broeck, Gioannini, Gon{\c{c}}alves, Quaggiotto, Colizza, and Vespignani]{Broeck2011}
Broeck, W. V.~d., Gioannini, C., Gon{\c{c}}alves, B., Quaggiotto, M., Colizza, V., and Vespignani, A.
\newblock The gleamviz computational tool, a publicly available software to explore realistic epidemic spreading scenarios at the global scale.
\newblock \emph{BMC Infectious Diseases}, 11\penalty0 (1):\penalty0 37, Feb 2011.
\newblock ISSN 1471-2334.
\newblock \doi{10.1186/1471-2334-11-37}.
\newblock URL \url{https://doi.org/10.1186/1471-2334-11-37}.

\bibitem[Brookes \& Roscoe(1984)Brookes and Roscoe]{brookes1984failures}
Brookes, S.~D. and Roscoe, A.~W.
\newblock An improved failures model for communicating processes.
\newblock In \emph{International Conference on Concurrency. Berlin, Heidelberg}. Springer, 1984.

\bibitem[Brookes et~al.(1984)Brookes, Hoare, and Roscoe]{brookes1984comseq}
Brookes, S.~D., Hoare, C.~A., and Roscoe, A.~W.
\newblock A theory of communicating sequential processes.
\newblock \emph{Journal of the ACM}, 31\penalty0 (3):\penalty0 560--599, 1984.

\bibitem[Brundage et~al.(2018)Brundage, Avin, Clark, Toner, Eckersley, Garfinkel, Dafoe, Scharre, Zeitzoff, Filar, Anderson, Roff, Allen, Steinhardt, Flynn, hEigeartaigh, Beard, Belfield, Farquhar, Lyle, Crootof, Evans, Page, Bryson, Yampolskiy, and Amodei]{brundage2018malicious}
Brundage, M., Avin, S., Clark, J., Toner, H., Eckersley, P., Garfinkel, B., Dafoe, A., Scharre, P., Zeitzoff, T., Filar, B., Anderson, H., Roff, H., Allen, G.~C., Steinhardt, J., Flynn, C., hEigeartaigh, S.~O., Beard, S., Belfield, H., Farquhar, S., Lyle, C., Crootof, R., Evans, O., Page, M., Bryson, J., Yampolskiy, R., and Amodei, D.
\newblock The malicious use of artificial intelligence: Forecasting, prevention, and mitigation, 2018.

\bibitem[Brynjolfsson \& Ng(2023)Brynjolfsson and Ng]{brynjolfsson2023big}
Brynjolfsson, E. and Ng, A.
\newblock Big ai can centralize decision-making and power, and that’s a problem.
\newblock \emph{Missing links in ai governance}, 65, 2023.

\bibitem[Buolamwini \& Gebru(2018)Buolamwini and Gebru]{buolamwini2018gender}
Buolamwini, J. and Gebru, T.
\newblock Gender shades: Intersectional accuracy disparities in commercial gender classification.
\newblock In \emph{Conference on fairness, accountability and transparency}, pp.\  77--91. PMLR, 2018.

\bibitem[Burgess(2023)]{jailbreaking}
Burgess, M.
\newblock The hacking of chatgpt is just getting started, Apr 2023.
\newblock URL \url{wired.com/story/chatgpt-jailbreak-generative-ai-hacking/}.
\newblock [Online; posted 12-April-2023].

\bibitem[Butler(1863)]{butler1863}
Butler, S.
\newblock Darwin among the machines.
\newblock 1863.

\bibitem[Cao et~al.(2021)Cao, Cohen, and Szpruch]{cao2021}
Cao, H., Cohen, S.~N., and Szpruch, L.
\newblock Identifiability in inverse reinforcement learning.
\newblock \emph{arXiv preprint}, arXiv:2106.03498 [cs.LG], 2021.

\bibitem[Carroll et~al.(2022)Carroll, Dragan, Russell, and Hadfield-Menell]{carroll2022estimating}
Carroll, M., Dragan, A., Russell, S., and Hadfield-Menell, D.
\newblock Estimating and penalizing induced preference shifts in recommender systems, 2022.

\bibitem[Carroll(2021)]{carroll2021quantum}
Carroll, S.~M.
\newblock The quantum field theory on which the everyday world supervenes, 2021.

\bibitem[Casper et~al.(2023)Casper, Davies, Shi, Gilbert, Scheurer, Rando, Freedman, Korbak, Lindner, Freire, Wang, Marks, Segerie, Carroll, Peng, Christoffersen, Damani, Slocum, Anwar, Siththaranjan, Nadeau, Michaud, Pfau, Krasheninnikov, Chen, Langosco, Hase, Bıyık, Dragan, Krueger, Sadigh, and Hadfield-Menell]{casper2023open}
Casper, S., Davies, X., Shi, C., Gilbert, T.~K., Scheurer, J., Rando, J., Freedman, R., Korbak, T., Lindner, D., Freire, P., Wang, T., Marks, S., Segerie, C.-R., Carroll, M., Peng, A., Christoffersen, P., Damani, M., Slocum, S., Anwar, U., Siththaranjan, A., Nadeau, M., Michaud, E.~J., Pfau, J., Krasheninnikov, D., Chen, X., Langosco, L., Hase, P., Bıyık, E., Dragan, A., Krueger, D., Sadigh, D., and Hadfield-Menell, D.
\newblock Open problems and fundamental limitations of reinforcement learning from human feedback, 2023.

\bibitem[Censi et~al.(2019)Censi, Slutsky, Wongpiromsarn, Yershov, Pendleton, Fu, and Frazzoli]{censi-icra19}
Censi, A., Slutsky, K., Wongpiromsarn, T., Yershov, D., Pendleton, S., Fu, J., and Frazzoli, E.
\newblock Liability, ethics, and culture-aware behavior specification using rulebooks.
\newblock In \emph{2019 International Conference on Robotics and Automation (ICRA)}, pp.\  8536--8542. IEEE, 2019.

\bibitem[Cohen et~al.(2022)Cohen, Hutter, and Osborne]{cohen2022advanced}
Cohen, M., Hutter, M., and Osborne, M.
\newblock Advanced artificial agents intervene in the provision of reward.
\newblock \emph{AI magazine}, 43\penalty0 (3):\penalty0 282--293, 2022.

\bibitem[Cook(2014)]{Cook2014BridgeFR}
Cook, W.
\newblock Bridge failure rates, consequences, and predictive trends.
\newblock 2014.
\newblock URL \url{https://api.semanticscholar.org/CorpusID:107360532}.

\bibitem[Crawford(2021)]{crawford2021atlas}
Crawford, K.
\newblock \emph{The atlas of AI: Power, politics, and the planetary costs of artificial intelligence}.
\newblock Yale University Press, 2021.

\bibitem[Cusumano-Towner et~al.(2019)Cusumano-Towner, Saad, Lew, and Mansinghka]{cusumano2019gen}
Cusumano-Towner, M.~F., Saad, F.~A., Lew, A.~K., and Mansinghka, V.~K.
\newblock Gen: a general-purpose probabilistic programming system with programmable inference.
\newblock In \emph{Proceedings of the 40th acm sigplan conference on programming language design and implementation}, pp.\  221--236, 2019.

\bibitem[Das et~al.(2023)Das, Stanton, and Wallace]{fairness_review_1}
Das, S., Stanton, R., and Wallace, N.
\newblock Algorithmic fairness.
\newblock \emph{Annual Review of Financial Economics}, 15\penalty0 (Volume 15, 2023):\penalty0 565--593, 2023.
\newblock ISSN 1941-1375.
\newblock \doi{https://doi.org/10.1146/annurev-financial-110921-125930}.
\newblock URL \url{https://www.annualreviews.org/content/journals/10.1146/annurev-financial-110921-125930}.

\bibitem['davidad' Dalrymple(2024)]{dalrymple2024}
'davidad' Dalrymple, D.
\newblock Safeguarded ai: constructing guaranteed safety.
\newblock Technical report, ARIA, feb 2024.
\newblock URL \url{https://www.aria.org.uk/wp-content/uploads/2024/01/ARIA-Safeguarded-AI-Programme-Thesis-V1.pdf}.

\bibitem[Del~Moral et~al.(2006)Del~Moral, Doucet, and Jasra]{del2006sequential}
Del~Moral, P., Doucet, A., and Jasra, A.
\newblock Sequential monte carlo samplers.
\newblock \emph{Journal of the Royal Statistical Society Series B: Statistical Methodology}, 68\penalty0 (3):\penalty0 411--436, 2006.

\bibitem[Deleu et~al.(2022)Deleu, Góis, Emezue, Rankawat, Lacoste-Julien, Bauer, and Bengio]{deleu2022bayesian}
Deleu, T., Góis, A., Emezue, C., Rankawat, M., Lacoste-Julien, S., Bauer, S., and Bengio, Y.
\newblock Bayesian structure learning with generative flow networks, 2022.

\bibitem[Deleu et~al.(2023)Deleu, Nishikawa-Toomey, Subramanian, Malkin, Charlin, and Bengio]{deleu2023joint}
Deleu, T., Nishikawa-Toomey, M., Subramanian, J., Malkin, N., Charlin, L., and Bengio, Y.
\newblock Joint bayesian inference of graphical structure and parameters with a single generative flow network, 2023.

\bibitem[Dhillon(2003)]{dhillon2003engineering}
Dhillon, B.
\newblock \emph{Engineering Safety: Fundamentals, Techniques, And Applications}.
\newblock Series On Industrial And Systems Engineering. World Scientific Publishing Company, 2003.
\newblock ISBN 9789813102361.
\newblock URL \url{https://books.google.co.uk/books?id=P_E7DQAAQBAJ}.

\bibitem[Dreossi et~al.(2018)Dreossi, Ghosh, Yue, Keutzer, Sangiovanni-Vincentelli, and Seshia]{dreossi-ijcai18}
Dreossi, T., Ghosh, S., Yue, X., Keutzer, K., Sangiovanni-Vincentelli, A., and Seshia, S.~A.
\newblock Counterexample-guided data augmentation.
\newblock In \emph{27th International Joint Conference on Artificial Intelligence (IJCAI)}, 2018.

\bibitem[Dreossi et~al.(2019)Dreossi, Donz{\'{e}}, and Seshia]{dreossi-jar19}
Dreossi, T., Donz{\'{e}}, A., and Seshia, S.~A.
\newblock Compositional falsification of cyber-physical systems with machine learning components.
\newblock \emph{Journal of Automated Reasoning}, 63\penalty0 (4):\penalty0 1031--1053, 2019.

\bibitem[Drexler(2023)]{drexler23}
Drexler, E.
\newblock The open agency model, February 2023.
\newblock URL \url{https://www.alignmentforum.org/posts/5hApNw5f7uG8RXxGS/the-open-agency-model}.
\newblock [Online; posted 22-February-2023].

\bibitem[Dvijotham \& Todorov(2010)Dvijotham and Todorov]{dvijotham2010}
Dvijotham, K. and Todorov, E.
\newblock Inverse optimal control with linearly-solvable {MDPs}.
\newblock In \emph{Proceedings of the 27th International Conference on Machine Learning}, pp.\  335--342, Haifa, Israel, June 2010. Omnipress, Madison, Wisconsin, USA.

\bibitem[Edwards et~al.(2023)Edwards, Peruffo, and Abate]{edwards2023general}
Edwards, A., Peruffo, A., and Abate, A.
\newblock A general verification framework for dynamical and control models via certificate synthesis, 2023.

\bibitem[Ellis et~al.(2023)Ellis, Wong, Nye, Sable-Meyer, Cary, Anaya~Pozo, Hewitt, Solar-Lezama, and Tenenbaum]{ellis2023dreamcoder}
Ellis, K., Wong, L., Nye, M., Sable-Meyer, M., Cary, L., Anaya~Pozo, L., Hewitt, L., Solar-Lezama, A., and Tenenbaum, J.~B.
\newblock Dreamcoder: growing generalizable, interpretable knowledge with wake--sleep bayesian program learning.
\newblock \emph{Philosophical Transactions of the Royal Society A}, 381\penalty0 (2251):\penalty0 20220050, 2023.

\bibitem[Elmaaroufi et~al.(2024)Elmaaroufi, Shankar, Cismaru, Vazquez-Chanlatte, Sangiovanni-Vincentelli, Zaharia, and Seshia]{elmaaroufi-arxiv24}
Elmaaroufi, K., Shankar, D., Cismaru, A., Vazquez-Chanlatte, M., Sangiovanni-Vincentelli, A., Zaharia, M., and Seshia, S.~A.
\newblock Generating probabilistic scenario programs from natural language.
\newblock \emph{ArXiV e-Prints 2405.03709}, 2024.

\bibitem[Ericson(2015)]{ericson2015hazard}
Ericson, C.
\newblock \emph{Hazard Analysis Techniques for System Safety}.
\newblock Wiley, 2015.
\newblock ISBN 9781118940389.
\newblock URL \url{https://books.google.co.uk/books?id=cTikBgAAQBAJ}.

\bibitem[Everitt et~al.(2021)Everitt, Carey, Langlois, Ortega, and Legg]{everitt2021agent}
Everitt, T., Carey, R., Langlois, E., Ortega, P.~A., and Legg, S.
\newblock Agent incentives: A causal perspective, 2021.

\bibitem[Feldstein(2019)]{feldstein2019global}
Feldstein, S.
\newblock \emph{The global expansion of AI surveillance}, volume~17.
\newblock Carnegie Endowment for International Peace Washington, DC, 2019.

\bibitem[Fikes \& Nilsson(1971)Fikes and Nilsson]{fikes1971strips}
Fikes, R.~E. and Nilsson, N.~J.
\newblock Strips: A new approach to the application of theorem proving to problem solving.
\newblock \emph{Artificial intelligence}, 2\penalty0 (3-4):\penalty0 189--208, 1971.

\bibitem[First et~al.(2023)First, Rabe, Ringer, and Brun]{first2023baldur}
First, E., Rabe, M., Ringer, T., and Brun, Y.
\newblock Baldur: Whole-proof generation and repair with large language models.
\newblock In \emph{Proceedings of the 31st ACM Joint European Software Engineering Conference and Symposium on the Foundations of Software Engineering}, pp.\  1229--1241, 2023.

\bibitem[FLI(2023)]{fli_safety_levels}
FLI.
\newblock {AI} governance scorecard and safety standards policy, 2023.
\newblock URL \url{https://futureoflife.org/wp-content/uploads/2023/11/FLI_Governance_Scorecard_and_Framework.pdf}.

\bibitem[Freedman et~al.(2020)Freedman, Shah, and Dragan]{choicesetmisspecification}
Freedman, R., Shah, R., and Dragan, A.
\newblock Choice set misspecification in reward inference.
\newblock In \emph{IJCAI-PRICAI-20 Workshop on Artificial Intelligence Safety}, 2020.
\newblock \doi{10.48550/ARXIV.2101.07691}.
\newblock URL \url{https://arxiv.org/abs/2101.07691}.

\bibitem[Fremont et~al.(2019)Fremont, Dreossi, Ghosh, Yue, Sangiovanni-Vincentelli, and Seshia]{fremont-pldi19}
Fremont, D.~J., Dreossi, T., Ghosh, S., Yue, X., Sangiovanni-Vincentelli, A.~L., and Seshia, S.~A.
\newblock Scenic: A language for scenario specification and scene generation.
\newblock In \emph{Proceedings of the 40th annual ACM SIGPLAN conference on Programming Language Design and Implementation (PLDI)}, June 2019.

\bibitem[Fremont et~al.(2020{\natexlab{a}})Fremont, Chiu, Margineantu, Osipychev, and Seshia]{fremont-cav20}
Fremont, D.~J., Chiu, J., Margineantu, D.~D., Osipychev, D., and Seshia, S.~A.
\newblock Formal analysis and redesign of a neural network-based aircraft taxiing system with {VerifAI}.
\newblock In \emph{32nd International Conference on Computer Aided Verification (CAV)}, July 2020{\natexlab{a}}.

\bibitem[Fremont et~al.(2020{\natexlab{b}})Fremont, Kim, Pant, Seshia, Acharya, Bruso, Wells, Lemke, Lu, and Mehta]{fremont-itsc20}
Fremont, D.~J., Kim, E., Pant, Y.~V., Seshia, S.~A., Acharya, A., Bruso, X., Wells, P., Lemke, S., Lu, Q., and Mehta, S.
\newblock Formal scenario-based testing of autonomous vehicles: From simulation to the real world.
\newblock In \emph{23rd {IEEE} International Conference on Intelligent Transportation Systems (ITSC)}, September 2020{\natexlab{b}}.

\bibitem[Fremont et~al.(2022)Fremont, Kim, Dreossi, Ghosh, Yue, Sangiovanni-Vincentelli, and Seshia]{scenic-mlj22}
Fremont, D.~J., Kim, E., Dreossi, T., Ghosh, S., Yue, X., Sangiovanni-Vincentelli, A.~L., and Seshia, S.~A.
\newblock Scenic: A language for scenario specification and data generation.
\newblock \emph{Machine Learning Journal}, 2022.

\bibitem[Ganguli et~al.(2022)Ganguli, Lovitt, Kernion, Askell, Bai, Kadavath, Mann, Perez, Schiefer, Ndousse, Jones, Bowman, Chen, Conerly, DasSarma, Drain, Elhage, El-Showk, Fort, Hatfield-Dodds, Henighan, Hernandez, Hume, Jacobson, Johnston, Kravec, Olsson, Ringer, Tran-Johnson, Amodei, Brown, Joseph, McCandlish, Olah, Kaplan, and Clark]{ganguli2022red}
Ganguli, D., Lovitt, L., Kernion, J., Askell, A., Bai, Y., Kadavath, S., Mann, B., Perez, E., Schiefer, N., Ndousse, K., Jones, A., Bowman, S., Chen, A., Conerly, T., DasSarma, N., Drain, D., Elhage, N., El-Showk, S., Fort, S., Hatfield-Dodds, Z., Henighan, T., Hernandez, D., Hume, T., Jacobson, J., Johnston, S., Kravec, S., Olsson, C., Ringer, S., Tran-Johnson, E., Amodei, D., Brown, T., Joseph, N., McCandlish, S., Olah, C., Kaplan, J., and Clark, J.
\newblock Red teaming language models to reduce harms: Methods, scaling behaviors, and lessons learned, 2022.

\bibitem[Gao \& Guan(2023)Gao and Guan]{gao2023interpretability}
Gao, L. and Guan, L.
\newblock Interpretability of machine learning: Recent advances and future prospects, 2023.

\bibitem[Garion et~al.(2022)Garion, Hattenberger, Pollien, Roux, and Thirioux]{autopilot_verification}
Garion, C., Hattenberger, G., Pollien, B., Roux, P., and Thirioux, X.
\newblock {Formal Verification for Autopilot - Preliminary state of the art}.
\newblock Technical report, {ISAE-SUPAERO ; ONERA -- The French Aerospace Lab ; ENAC}, March 2022.
\newblock URL \url{https://hal.science/hal-03255656}.

\bibitem[Ghallab et~al.(2004)Ghallab, Nau, and Traverso]{ghallab2004automated}
Ghallab, M., Nau, D., and Traverso, P.
\newblock \emph{Automated Planning: theory and practice}.
\newblock Elsevier, 2004.

\bibitem[Good(1959)]{good1959}
Good, I.~J.
\newblock Speculations on perceptrons and other automata.
\newblock \emph{International Business Machines Corporation}, 1959.

\bibitem[Goodhart(1975)]{goodhart_problems_1975}
Goodhart, C.
\newblock Problems of monetary management: the {UK} experience in papers in monetary economics.
\newblock \emph{Monetary Economics}, 1, 1975.

\bibitem[Goodman et~al.(2012)Goodman, Mansinghka, Roy, Bonawitz, and Tenenbaum]{goodman2012church}
Goodman, N., Mansinghka, V., Roy, D.~M., Bonawitz, K., and Tenenbaum, J.~B.
\newblock Church: a language for generative models.
\newblock \emph{arXiv preprint arXiv:1206.3255}, 2012.

\bibitem[Gothoskar et~al.(2021)Gothoskar, Cusumano-Towner, Zinberg, Ghavamizadeh, Pollok, Garrett, Tenenbaum, Gutfreund, and Mansinghka]{gothoskar20213dp3}
Gothoskar, N., Cusumano-Towner, M., Zinberg, B., Ghavamizadeh, M., Pollok, F., Garrett, A., Tenenbaum, J., Gutfreund, D., and Mansinghka, V.
\newblock 3dp3: 3d scene perception via probabilistic programming.
\newblock \emph{Advances in Neural Information Processing Systems}, 34:\penalty0 9600--9612, 2021.

\bibitem[Gothoskar et~al.(2023)Gothoskar, Ghavami, Li, Curtis, Noseworthy, Chung, Patton, Freeman, Tenenbaum, Klukas, and Mansinghka]{gothoskar2023bayes3d}
Gothoskar, N., Ghavami, M., Li, E., Curtis, A., Noseworthy, M., Chung, K., Patton, B., Freeman, W.~T., Tenenbaum, J.~B., Klukas, M., and Mansinghka, V.~K.
\newblock Bayes3d: fast learning and inference in structured generative models of 3d objects and scenes, 2023.

\bibitem[Grand et~al.(2023)Grand, Wong, Bowers, Olausson, Liu, Tenenbaum, and Andreas]{grand2023lilo}
Grand, G., Wong, L., Bowers, M., Olausson, T.~X., Liu, M., Tenenbaum, J.~B., and Andreas, J.
\newblock Lilo: Learning interpretable libraries by compressing and documenting code.
\newblock In \emph{The Twelfth International Conference on Learning Representations}, 2023.

\bibitem[Gulwani et~al.(2017)Gulwani, Polozov, and Singh]{PGL-010}
Gulwani, S., Polozov, O., and Singh, R.
\newblock Program synthesis.
\newblock \emph{Foundations and Trends® in Programming Languages}, 4\penalty0 (1-2):\penalty0 1--119, 2017.
\newblock ISSN 2325-1107.
\newblock \doi{10.1561/2500000010}.
\newblock URL \url{http://dx.doi.org/10.1561/2500000010}.

\bibitem[Guo et~al.(2021)Guo, Tondi, and Barni]{guo2021overview}
Guo, W., Tondi, B., and Barni, M.
\newblock An overview of backdoor attacks against deep neural networks and possible defences, 2021.

\bibitem[Hadfield-Menell et~al.(2017)Hadfield-Menell, Dragan, Abbeel, and Russell]{hadfieldmenell2017offswitch}
Hadfield-Menell, D., Dragan, A., Abbeel, P., and Russell, S.
\newblock The off-switch game, 2017.

\bibitem[Hadfield-Menell et~al.(2024)Hadfield-Menell, Dragan, Abbeel, and Russell]{hadfieldmenell2024cooperative}
Hadfield-Menell, D., Dragan, A., Abbeel, P., and Russell, S.
\newblock Cooperative inverse reinforcement learning, 2024.

\bibitem[Halpern(2016)]{Halpern16}
Halpern, J.~Y.
\newblock \emph{Actual Causality}.
\newblock MIT Press, Cambridge, MA, 2016.
\newblock ISBN 978-0-262-03502-6.
\newblock \doi{10.7551/mitpress/9780262035026.001.0001}.

\bibitem[Halpern \& Leung(2013)Halpern and Leung]{halpern2013}
Halpern, J.~Y. and Leung, S.
\newblock Weighted sets of probabilities and minimax weighted expected regret: new approaches for representing uncertainty and making decisions.
\newblock \emph{arXiv preprint arXiv:1302.5681}, 2013.

\bibitem[Hansson \& Jonsson(1994)Hansson and Jonsson]{PCTL}
Hansson, H. and Jonsson, B.
\newblock A logic for reasoning about time and reliability.
\newblock \emph{Form. Asp. Comput.}, 6\penalty0 (5):\penalty0 512–535, sep 1994.
\newblock ISSN 0934-5043.
\newblock \doi{10.1007/BF01211866}.
\newblock URL \url{https://doi.org/10.1007/BF01211866}.

\bibitem[Hasanbeig et~al.(2023)Hasanbeig, Kroening, and Abate]{Hasanbeig2023-HASCRL}
Hasanbeig, H., Kroening, D., and Abate, A.
\newblock Certified reinforcement learning with logic guidance.
\newblock \emph{Artificial Intelligence}, 322\penalty0 (C):\penalty0 103949, 2023.
\newblock \doi{10.1016/j.artint.2023.103949}.

\bibitem[Hendrycks et~al.(2023)Hendrycks, Mazeika, and Woodside]{hendrycks2023overview}
Hendrycks, D., Mazeika, M., and Woodside, T.
\newblock An overview of catastrophic ai risks, 2023.

\bibitem[Hennessy \& Goodhart(2023)Hennessy and Goodhart]{hennessy_goodharts_2023}
Hennessy, C.~A. and Goodhart, C. A.~E.
\newblock Goodhart's {Law} and {Machine} {Learning}: {A} {Structural} {Perspective}.
\newblock \emph{International Economic Review}, pp.\  iere.12633, March 2023.
\newblock ISSN 0020-6598, 1468-2354.
\newblock \doi{10.1111/iere.12633}.
\newblock URL \url{https://onlinelibrary.wiley.com/doi/10.1111/iere.12633}.

\bibitem[Hinton(2023)]{interview_hinton}
Hinton, G.
\newblock ‘the godfather of {A.I.}’ leaves {Google} and warns of danger ahead, May 2023.
\newblock URL \url{https://www.nytimes.com/2023/05/01/technology/ai-google-chatbot-engineer-quits-hinton.html}.
\newblock [Online; posted 4-May-2023].

\bibitem[Hollmann et~al.(2023)Hollmann, Müller, Eggensperger, and Hutter]{hollmann2023tabpfn}
Hollmann, N., Müller, S., Eggensperger, K., and Hutter, F.
\newblock {TabPFN}: A transformer that solves small tabular classification problems in a second, 2023.

\bibitem[Hu et~al.(2023)Hu, Malkin, Jain, Everett, Graikos, and Bengio]{hu2023gflownetem}
Hu, E.~J., Malkin, N., Jain, M., Everett, K., Graikos, A., and Bengio, Y.
\newblock Gflownet-em for learning compositional latent variable models, 2023.

\bibitem[Hubinger et~al.(2021)Hubinger, van Merwijk, Mikulik, Skalse, and Garrabrant]{hubinger2021risks}
Hubinger, E., van Merwijk, C., Mikulik, V., Skalse, J., and Garrabrant, S.
\newblock Risks from learned optimization in advanced machine learning systems, 2021.

\bibitem[Hutter(2003)]{hutter2003gentle}
Hutter, M.
\newblock A gentle introduction to the universal algorithmic agent aixi, 2003.

\bibitem[Ivanov et~al.(2020)Ivanov, Carpenter, Weimer, Alur, Pappas, and Lee]{car_verification_case_study}
Ivanov, R., Carpenter, T.~J., Weimer, J., Alur, R., Pappas, G.~J., and Lee, I.
\newblock Case study: verifying the safety of an autonomous racing car with a neural network controller.
\newblock In \emph{Proceedings of the 23rd International Conference on Hybrid Systems: Computation and Control}, HSCC '20, New York, NY, USA, 2020. Association for Computing Machinery.
\newblock ISBN 9781450370189.
\newblock \doi{10.1145/3365365.3382216}.
\newblock URL \url{https://doi.org/10.1145/3365365.3382216}.

\bibitem[Jacob et~al.(2023)Jacob, Shen, Farina, and Andreas]{jacob2023consensus}
Jacob, A.~P., Shen, Y., Farina, G., and Andreas, J.
\newblock The consensus game: Language model generation via equilibrium search.
\newblock \emph{arXiv preprint arXiv:2310.09139}, 2023.

\bibitem[Jenner \& Gleave(2022)Jenner and Gleave]{jenner2022preprocessing}
Jenner, E. and Gleave, A.
\newblock Preprocessing reward functions for interpretability, 2022.

\bibitem[{Jha} \& {Seshia}(2017){Jha} and {Seshia}]{jha-acta17}
{Jha}, S. and {Seshia}, S.~A.
\newblock {A Theory of Formal Synthesis via Inductive Learning}.
\newblock \emph{Acta Informatica}, 54\penalty0 (7):\penalty0 693--726, 2017.

\bibitem[Ji et~al.(2024)Ji, Qiu, Chen, Zhang, Lou, Wang, Duan, He, Zhou, Zhang, Zeng, Ng, Dai, Pan, O'Gara, Lei, Xu, Tse, Fu, McAleer, Yang, Wang, Zhu, Guo, and Gao]{ji2024ai}
Ji, J., Qiu, T., Chen, B., Zhang, B., Lou, H., Wang, K., Duan, Y., He, Z., Zhou, J., Zhang, Z., Zeng, F., Ng, K.~Y., Dai, J., Pan, X., O'Gara, A., Lei, Y., Xu, H., Tse, B., Fu, J., McAleer, S., Yang, Y., Wang, Y., Zhu, S.-C., Guo, Y., and Gao, W.
\newblock Ai alignment: A comprehensive survey, 2024.

\bibitem[Joy(2000)]{joy2000}
Joy, B.
\newblock Why the future doesn't need us, 2000.

\bibitem[Junges et~al.(2016)Junges, Jansen, Dehnert, Topcu, and Katoen]{Junges2016}
Junges, S., Jansen, N., Dehnert, C., Topcu, U., and Katoen, J.-P.
\newblock Safety-constrained reinforcement learning for {MDP}s.
\newblock In Chechik, M. and Raskin, J.-F. (eds.), \emph{Tools and Algorithms for the Construction and Analysis of Systems}, pp.\  130--146, Berlin, Heidelberg, 2016. Springer Berlin Heidelberg.
\newblock ISBN 978-3-662-49674-9.

\bibitem[Karwowski et~al.(2023)Karwowski, Hayman, Bai, Kiendlhofer, Griffin, and Skalse]{karwowski2023goodharts}
Karwowski, J., Hayman, O., Bai, X., Kiendlhofer, K., Griffin, C., and Skalse, J.
\newblock Goodhart's law in reinforcement learning, 2023.

\bibitem[Ke et~al.(2022)Ke, Chiappa, Wang, Goyal, Bornschein, Rey, Weber, Botvinic, Mozer, and Rezende]{ke2022learning}
Ke, N.~R., Chiappa, S., Wang, J., Goyal, A., Bornschein, J., Rey, M., Weber, T., Botvinic, M., Mozer, M., and Rezende, D.~J.
\newblock Learning to induce causal structure, 2022.

\bibitem[Kearns \& Vazirani(1994)Kearns and Vazirani]{introtoCLT}
Kearns, M.~J. and Vazirani, U.
\newblock \emph{{An Introduction to Computational Learning Theory}}.
\newblock The MIT Press, 08 1994.
\newblock ISBN 9780262276863.
\newblock \doi{10.7551/mitpress/3897.001.0001}.
\newblock URL \url{https://doi.org/10.7551/mitpress/3897.001.0001}.

\bibitem[Kemp et~al.(2010)Kemp, Tenenbaum, Niyogi, and Griffiths]{kemp2010probabilistic}
Kemp, C., Tenenbaum, J.~B., Niyogi, S., and Griffiths, T.~L.
\newblock A probabilistic model of theory formation.
\newblock \emph{Cognition}, 114\penalty0 (2):\penalty0 165--196, 2010.

\bibitem[Khlaaf(2023)]{khlaaf2023toward}
Khlaaf, H.
\newblock Toward comprehensive risk assessments and assurance of ai-based systems.
\newblock \emph{Trail of Bits}, 2023.

\bibitem[Khlaaf et~al.(2022)Khlaaf, Mishkin, Achiam, Krueger, and Brundage]{khlaaf2022hazard}
Khlaaf, H., Mishkin, P., Achiam, J., Krueger, G., and Brundage, M.
\newblock A hazard analysis framework for code synthesis large language models, 2022.

\bibitem[Kim et~al.(2021)Kim, Garg, Shiragur, and Ermon]{kim2021}
Kim, K., Garg, S., Shiragur, K., and Ermon, S.
\newblock Reward identification in inverse reinforcement learning.
\newblock In \emph{Proceedings of the 38th International Conference on Machine Learning}, volume 139 of \emph{Proceedings of Machine Learning Research}, pp.\  5496--5505, Virtual, July 2021. PMLR.

\bibitem[Kleinberg et~al.(2016)Kleinberg, Mullainathan, and Raghavan]{kleinberg2016inherent}
Kleinberg, J., Mullainathan, S., and Raghavan, M.
\newblock Inherent trade-offs in the fair determination of risk scores, 2016.

\bibitem[Koessler \& Schuett(2023)Koessler and Schuett]{koessler2023risk}
Koessler, L. and Schuett, J.
\newblock Risk assessment at agi companies: A review of popular risk assessment techniques from other safety-critical industries, 2023.

\bibitem[Kosoy(2021)]{kosoy2021infrabayes}
Kosoy, V.
\newblock Infra-bayesian physicalism: a formal theory of naturalized induction, November 2021.
\newblock URL \url{https://www.alignmentforum.org/posts/gHgs2e2J5azvGFatb/infra-bayesian-physicalism-a-formal-theory-of-naturalized}.
\newblock [Online; posted 30-November-2021].

\bibitem[Krakovna et~al.(2020)Krakovna, Uesato, Mikulik, Rahtz, Everitt, Kumar, Kenton, Leike, and Legg]{krakovna_specification_2020}
Krakovna, V., Uesato, J., Mikulik, V., Rahtz, M., Everitt, T., Kumar, R., Kenton, Z., Leike, J., and Legg, S.
\newblock Specification gaming: the flip side of {AI} ingenuity, 2020.
\newblock URL \url{deepmind.com/blog/specification-gaming-the-flip-side-of-ai-ingenuity}.

\bibitem[Lample et~al.(2017)Lample, Conneau, Denoyer, and Ranzato]{lample2017unsupervised}
Lample, G., Conneau, A., Denoyer, L., and Ranzato, M.
\newblock Unsupervised machine translation using monolingual corpora only.
\newblock \emph{arXiv preprint arXiv:1711.00043}, 2017.

\bibitem[Lample et~al.(2022)Lample, Lachaux, Lavril, Martinet, Hayat, Ebner, Rodriguez, and Lacroix]{lample2022hypertree}
Lample, G., Lachaux, M.-A., Lavril, T., Martinet, X., Hayat, A., Ebner, G., Rodriguez, A., and Lacroix, T.
\newblock Hypertree proof search for neural theorem proving, 2022.

\bibitem[Langosco et~al.(2023)Langosco, Koch, Sharkey, Pfau, Orseau, and Krueger]{langosco2023goal}
Langosco, L., Koch, J., Sharkey, L., Pfau, J., Orseau, L., and Krueger, D.
\newblock Goal misgeneralization in deep reinforcement learning, 2023.

\bibitem[LaValle \& Hutchinson(1996)LaValle and Hutchinson]{lavalle1996}
LaValle, S.~M. and Hutchinson, S.~A.
\newblock Evaluating motion strategies under nondeterministic or probabilistic uncertainties in sensing and control.
\newblock \emph{Proceedings of IEEE International Conference on Robotics and Automation}, 4, 1996.

\bibitem[Lazar \& Nelson(2023)Lazar and Nelson]{lazar2023ai}
Lazar, S. and Nelson, A.
\newblock Ai safety on whose terms?, 2023.

\bibitem[Leino(2023)]{leino2023program}
Leino, K. R.~M.
\newblock \emph{Program Proofs}.
\newblock MIT Press, 2023.

\bibitem[Lelkes et~al.(2017)Lelkes, Sood, and Iyengar]{hostileaudience}
Lelkes, Y., Sood, G., and Iyengar, S.
\newblock The hostile audience: The effect of access to broadband internet on partisan affect.
\newblock \emph{American Journal of Political Science}, 61\penalty0 (1):\penalty0 5--20, 2017.
\newblock ISSN 00925853, 15405907.
\newblock URL \url{http://www.jstor.org/stable/26379489}.

\bibitem[Leveson(2012)]{engineering_a_safer_world}
Leveson, N.
\newblock \emph{Engineering a Safer World: Systems Thinking Applied to Safety}.
\newblock 01 2012.
\newblock ISBN 9780262298247.
\newblock \doi{10.7551/mitpress/8179.001.0001}.

\bibitem[Lew et~al.(2020)Lew, Tessler, Mansinghka, and Tenenbaum]{lew2020thoughtppl}
Lew, A.~K., Tessler, M.~H., Mansinghka, V.~K., and Tenenbaum, J.~B.
\newblock Leveraging unstructured statistical knowledge in a probabilistic language of thought.
\newblock In \emph{CogSci 2020: Proceedings of the Forty-Second Annual Virtual Meeting of the Cognitive Science Society}, Proceedings of the Annual Conference of the Cognitive Science Society, 2020.

\bibitem[Lew et~al.(2023{\natexlab{a}})Lew, Huot, Staton, and Mansinghka]{lew2023adev}
Lew, A.~K., Huot, M., Staton, S., and Mansinghka, V.~K.
\newblock Adev: Sound automatic differentiation of expected values of probabilistic programs.
\newblock \emph{Proceedings of the ACM on Programming Languages}, 7\penalty0 (POPL):\penalty0 121--153, 2023{\natexlab{a}}.

\bibitem[Lew et~al.(2023{\natexlab{b}})Lew, Matheos, Zhi-Xuan, Ghavamizadeh, Gothoskar, Russell, and Mansinghka]{lew2023smcp3}
Lew, A.~K., Matheos, G., Zhi-Xuan, T., Ghavamizadeh, M., Gothoskar, N., Russell, S., and Mansinghka, V.~K.
\newblock {SMCP3}: Sequential monte carlo with probabilistic program proposals.
\newblock In Ruiz, F., Dy, J., and van~de Meent, J.-W. (eds.), \emph{Proceedings of The 26th International Conference on Artificial Intelligence and Statistics}, volume 206 of \emph{Proceedings of Machine Learning Research}, pp.\  7061--7088. PMLR, 25--27 Apr 2023{\natexlab{b}}.
\newblock URL \url{https://proceedings.mlr.press/v206/lew23a.html}.

\bibitem[Liell-Cock \& Staton(2024)Liell-Cock and Staton]{liellcock2024}
Liell-Cock, J. and Staton, S.
\newblock Compositional imprecise probability.
\newblock \emph{arXiv preprint arXiv:2405.09391}, 2024.

\bibitem[Liu et~al.(2020)Liu, Du, Wang, and {Da Young}]{liu2020bully}
Liu, P., Du, Y., Wang, L., and {Da Young}, J.
\newblock Ready to bully automated vehicles on public roads?
\newblock \emph{Accident Analysis and Prevention}, 137, March 2020.
\newblock ISSN 0001-4575.
\newblock \doi{10.1016/j.aap.2020.105457}.
\newblock Publisher Copyright: {\textcopyright} 2020 Elsevier Ltd.

\bibitem[London \& Heidari(2023)London and Heidari]{london2023beneficent}
London, A.~J. and Heidari, H.
\newblock Beneficent intelligence: A capability approach to modeling benefit, assistance, and associated moral failures through ai systems, 2023.

\bibitem[Loveland(2016)]{loveland2016automated}
Loveland, D.~W.
\newblock \emph{Automated theorem proving: A logical basis}.
\newblock Elsevier, 2016.

\bibitem[Manheim \& Garrabrant(2019)Manheim and Garrabrant]{manheim_categorizing_2019}
Manheim, D. and Garrabrant, S.
\newblock Categorizing {Variants} of {Goodhart}'s {Law}, February 2019.
\newblock URL \url{http://arxiv.org/abs/1803.04585}.
\newblock arXiv:1803.04585 [cs, q-fin, stat].

\bibitem[Mao et~al.(2019)Mao, Gan, Kohli, Tenenbaum, and Wu]{mao2019neuro}
Mao, J., Gan, C., Kohli, P., Tenenbaum, J.~B., and Wu, J.
\newblock The neuro-symbolic concept learner: Interpreting scenes, words, and sentences from natural supervision.
\newblock \emph{arXiv preprint arXiv:1904.12584}, 2019.

\bibitem[Matheos et~al.(2020)Matheos, Lew, Ghavamizadeh, Russell, Cusumano-Towner, and Mansinghka]{matheos2020transforming}
Matheos, G., Lew, A.~K., Ghavamizadeh, M., Russell, S., Cusumano-Towner, M., and Mansinghka, V.
\newblock Transforming worlds: Automated involutive mcmc for open-universe probabilistic models.
\newblock In \emph{Third Symposium on Advances in Approximate Bayesian Inference}, 2020.

\bibitem[Mathews \& Schmidler(2022)Mathews and Schmidler]{mathews2022finite}
Mathews, J. and Schmidler, S.~C.
\newblock Finite sample complexity of sequential monte carlo estimators on multimodal target distributions, 2022.

\bibitem[McCullough(2001)]{mccullough2001great}
McCullough, D.
\newblock \emph{The Great Bridge: The Epic Story of the Building of the Brooklyn Bridge}.
\newblock Simon \& Schuster, 2001.
\newblock ISBN 9780743217378.
\newblock URL \url{https://books.google.co.uk/books?id=bOM93rb22YEC}.

\bibitem[McMahan et~al.(2005)McMahan, Likhachev, and Gordon]{McMahan2005}
McMahan, H.~B., Likhachev, M., and Gordon, G.~J.
\newblock Bounded real-time dynamic programming: Rtdp with monotone upper bounds and performance guarantees.
\newblock In \emph{Proceedings of the 22nd International Conference on Machine Learning}, ICML '05, pp.\  569–576, New York, NY, USA, 2005. Association for Computing Machinery.
\newblock ISBN 1595931805.
\newblock \doi{10.1145/1102351.1102423}.
\newblock URL \url{https://doi.org/10.1145/1102351.1102423}.

\bibitem[Megill(2023)]{Metamath}
Megill, N.
\newblock {Proof Explorer - Home Page - Metamath}, September 2023.
\newblock URL \url{https://us.metamath.org/mpeuni/mmset.html}.
\newblock [Online; accessed 3. Sep. 2023].

\bibitem[Memarian \& Doleck(2023)Memarian and Doleck]{FATEoverview}
Memarian, B. and Doleck, T.
\newblock Fairness, accountability, transparency, and ethics (fate) in artificial intelligence (ai) and higher education: A systematic review.
\newblock \emph{Computers and Education: Artificial Intelligence}, 5:\penalty0 100152, 2023.
\newblock ISSN 2666-920X.
\newblock \doi{https://doi.org/10.1016/j.caeai.2023.100152}.
\newblock URL \url{https://www.sciencedirect.com/science/article/pii/S2666920X23000310}.

\bibitem[Michaud et~al.(2020)Michaud, Gleave, and Russell]{michaud2020understanding}
Michaud, E.~J., Gleave, A., and Russell, S.
\newblock Understanding learned reward functions, 2020.

\bibitem[Michaud et~al.(2024)Michaud, Liao, Lad, Liu, Mudide, Loughridge, Guo, Kheirkhah, Vukeli{\'c}, and Tegmark]{michaud2024opening}
Michaud, E.~J., Liao, I., Lad, V., Liu, Z., Mudide, A., Loughridge, C., Guo, Z.~C., Kheirkhah, T.~R., Vukeli{\'c}, M., and Tegmark, M.
\newblock Opening the ai black box: program synthesis via mechanistic interpretability.
\newblock \emph{arXiv preprint arXiv:2402.05110}, 2024.

\bibitem[Milch et~al.(2007)Milch, Marthi, Russell, Sontag, Ong, and Kolobov]{blog}
Milch, B., Marthi, B., Russell, S., Sontag, D., Ong, D.~L., and Kolobov, A.
\newblock {Blog}: Probabilistic models with unknown objects.
\newblock \emph{Statistical Relational Learning}, pp.\  373, 2007.

\bibitem[Mingard et~al.(2020)Mingard, Skalse, Valle-Pérez, Martínez-Rubio, Mikulik, and Louis]{NNentropy}
Mingard, C., Skalse, J., Valle-Pérez, G., Martínez-Rubio, D., Mikulik, V., and Louis, A.~A.
\newblock Neural networks are a priori biased towards boolean functions with low entropy, 2020.

\bibitem[Mingard et~al.(2021)Mingard, Valle-Pérez, Skalse, and Louis]{SGDBayes}
Mingard, C., Valle-Pérez, G., Skalse, J., and Louis, A.~A.
\newblock Is sgd a bayesian sampler? well, almost.
\newblock \emph{Journal of Machine Learning Research}, 22\penalty0 (79):\penalty0 1--64, 2021.
\newblock URL \url{http://jmlr.org/papers/v22/20-676.html}.

\bibitem[Minsky(1986)]{minsky1984}
Minsky, M.
\newblock \emph{Afterword to Vernor Vinge’s novel,“True name}.
\newblock 1986.

\bibitem[Mio et~al.(2021)Mio, Sarkis, and Vignudelli]{mio21combiningnondet}
Mio, M., Sarkis, R., and Vignudelli, V.
\newblock Combining nondeterminism, probability, and termination: equational and metric reasoning.
\newblock \emph{36th Annual ACM/IEEE Symposium on Logic in Computer Science (LICS)}, 52/9:\penalty0 1--14, 2021.

\bibitem[Moravec(1988)]{moravec1988}
Moravec, H.
\newblock \emph{ind Children: The Future of Robot and Human Intelligence}.
\newblock Harvard University Press, 1988.

\bibitem[Mutlu \& Kim(2019)Mutlu and Kim]{mutlu2019rowhammer}
Mutlu, O. and Kim, J.~S.
\newblock Rowhammer: A retrospective.
\newblock \emph{IEEE Transactions on Computer-Aided Design of Integrated Circuits and Systems}, 39\penalty0 (8):\penalty0 1555--1571, 2019.

\bibitem[Neider \& Gavran(2018)Neider and Gavran]{neider2018learning}
Neider, D. and Gavran, I.
\newblock Learning linear temporal properties, 2018.

\bibitem[Newell \& Simon(1961)Newell and Simon]{gps1961}
Newell, A. and Simon, H.~A.
\newblock Gps, a program that simulates human thought.
\newblock 1961.

\bibitem[Ng \& Russell(2000)Ng and Russell]{ng2000}
Ng, A.~Y. and Russell, S.
\newblock Algorithms for inverse reinforcement learning.
\newblock In \emph{Proceedings of the Seventeenth International Conference on Machine Learning}, volume~1, pp.\  663--670, Stanford, California, USA, 2000. Morgan Kaufmann Publishers Inc.

\bibitem[Nieuwhof(1975)]{nieuwhof1975introduction}
Nieuwhof, G.
\newblock An introduction to fault tree analysis with emphasis on failure rate evaluation.
\newblock \emph{Microelectronics Reliability}, 14\penalty0 (2):\penalty0 105--119, 1975.
\newblock ISSN 0026-2714.
\newblock \doi{10.1016/0026-2714(75)90024-4}.
\newblock URL \url{https://www.sciencedirect.com/science/article/pii/0026271475900244}.

\bibitem[O'Keefe \& O'Leary(1993)O'Keefe and O'Leary]{o1993expert}
O'Keefe, R.~M. and O'Leary, D.~E.
\newblock Expert system verification and validation: a survey and tutorial.
\newblock \emph{Artificial Intelligence Review}, 7:\penalty0 3--42, 1993.

\bibitem[OpenAI et~al.(2024)OpenAI, :, Achiam, Adler, Agarwal, Ahmad, Akkaya, Aleman, Almeida, Altenschmidt, Altman, Anadkat, Avila, Babuschkin, Balaji, Balcom, Baltescu, Bao, Bavarian, Belgum, Bello, Berdine, Bernadett-Shapiro, Berner, Bogdonoff, Boiko, Boyd, Brakman, Brockman, Brooks, Brundage, Button, Cai, Campbell, Cann, Carey, Carlson, Carmichael, Chan, Chang, Chantzis, Chen, Chen, Chen, Chen, Chen, Chess, Cho, Chu, Chung, Cummings, Currier, Dai, Decareaux, Degry, Deutsch, Deville, Dhar, Dohan, Dowling, Dunning, Ecoffet, Eleti, Eloundou, Farhi, Fedus, Felix, Fishman, Forte, Fulford, Gao, Georges, Gibson, Goel, Gogineni, Goh, Gontijo-Lopes, Gordon, Grafstein, Gray, Greene, Gross, Gu, Guo, Hallacy, Han, Harris, He, Heaton, Heidecke, Hesse, Hickey, Hickey, Hoeschele, Houghton, Hsu, Hu, Hu, Huizinga, Jain, Jain, Jang, Jiang, Jiang, Jin, Jin, Jomoto, Jonn, Jun, Kaftan, Łukasz Kaiser, Kamali, Kanitscheider, Keskar, Khan, Kilpatrick, Kim, Kim, Kim, Kirchner, Kiros, Knight, Kokotajlo, Łukasz Kondraciuk,
  Kondrich, Konstantinidis, Kosic, Krueger, Kuo, Lampe, Lan, Lee, Leike, Leung, Levy, Li, Lim, Lin, Lin, Litwin, Lopez, Lowe, Lue, Makanju, Malfacini, Manning, Markov, Markovski, Martin, Mayer, Mayne, McGrew, McKinney, McLeavey, McMillan, McNeil, Medina, Mehta, Menick, Metz, Mishchenko, Mishkin, Monaco, Morikawa, Mossing, Mu, Murati, Murk, Mély, Nair, Nakano, Nayak, Neelakantan, Ngo, Noh, Ouyang, O'Keefe, Pachocki, Paino, Palermo, Pantuliano, Parascandolo, Parish, Parparita, Passos, Pavlov, Peng, Perelman, de~Avila Belbute~Peres, Petrov, de~Oliveira~Pinto, Michael, Pokorny, Pokrass, Pong, Powell, Power, Power, Proehl, Puri, Radford, Rae, Ramesh, Raymond, Real, Rimbach, Ross, Rotsted, Roussez, Ryder, Saltarelli, Sanders, Santurkar, Sastry, Schmidt, Schnurr, Schulman, Selsam, Sheppard, Sherbakov, Shieh, Shoker, Shyam, Sidor, Sigler, Simens, Sitkin, Slama, Sohl, Sokolowsky, Song, Staudacher, Such, Summers, Sutskever, Tang, Tezak, Thompson, Tillet, Tootoonchian, Tseng, Tuggle, Turley, Tworek, Uribe, Vallone,
  Vijayvergiya, Voss, Wainwright, Wang, Wang, Wang, Ward, Wei, Weinmann, Welihinda, Welinder, Weng, Weng, Wiethoff, Willner, Winter, Wolrich, Wong, Workman, Wu, Wu, Wu, Xiao, Xu, Yoo, Yu, Yuan, Zaremba, Zellers, Zhang, Zhang, Zhao, Zheng, Zhuang, Zhuk, and Zoph]{openai2024gpt4}
OpenAI, :, Achiam, J., Adler, S., Agarwal, S., Ahmad, L., Akkaya, I., Aleman, F.~L., Almeida, D., Altenschmidt, J., Altman, S., Anadkat, S., Avila, R., Babuschkin, I., Balaji, S., Balcom, V., Baltescu, P., Bao, H., Bavarian, M., Belgum, J., Bello, I., Berdine, J., Bernadett-Shapiro, G., Berner, C., Bogdonoff, L., Boiko, O., Boyd, M., Brakman, A.-L., Brockman, G., Brooks, T., Brundage, M., Button, K., Cai, T., Campbell, R., Cann, A., Carey, B., Carlson, C., Carmichael, R., Chan, B., Chang, C., Chantzis, F., Chen, D., Chen, S., Chen, R., Chen, J., Chen, M., Chess, B., Cho, C., Chu, C., Chung, H.~W., Cummings, D., Currier, J., Dai, Y., Decareaux, C., Degry, T., Deutsch, N., Deville, D., Dhar, A., Dohan, D., Dowling, S., Dunning, S., Ecoffet, A., Eleti, A., Eloundou, T., Farhi, D., Fedus, L., Felix, N., Fishman, S.~P., Forte, J., Fulford, I., Gao, L., Georges, E., Gibson, C., Goel, V., Gogineni, T., Goh, G., Gontijo-Lopes, R., Gordon, J., Grafstein, M., Gray, S., Greene, R., Gross, J., Gu, S.~S., Guo, Y.,
  Hallacy, C., Han, J., Harris, J., He, Y., Heaton, M., Heidecke, J., Hesse, C., Hickey, A., Hickey, W., Hoeschele, P., Houghton, B., Hsu, K., Hu, S., Hu, X., Huizinga, J., Jain, S., Jain, S., Jang, J., Jiang, A., Jiang, R., Jin, H., Jin, D., Jomoto, S., Jonn, B., Jun, H., Kaftan, T., Łukasz Kaiser, Kamali, A., Kanitscheider, I., Keskar, N.~S., Khan, T., Kilpatrick, L., Kim, J.~W., Kim, C., Kim, Y., Kirchner, J.~H., Kiros, J., Knight, M., Kokotajlo, D., Łukasz Kondraciuk, Kondrich, A., Konstantinidis, A., Kosic, K., Krueger, G., Kuo, V., Lampe, M., Lan, I., Lee, T., Leike, J., Leung, J., Levy, D., Li, C.~M., Lim, R., Lin, M., Lin, S., Litwin, M., Lopez, T., Lowe, R., Lue, P., Makanju, A., Malfacini, K., Manning, S., Markov, T., Markovski, Y., Martin, B., Mayer, K., Mayne, A., McGrew, B., McKinney, S.~M., McLeavey, C., McMillan, P., McNeil, J., Medina, D., Mehta, A., Menick, J., Metz, L., Mishchenko, A., Mishkin, P., Monaco, V., Morikawa, E., Mossing, D., Mu, T., Murati, M., Murk, O., Mély, D., Nair, A.,
  Nakano, R., Nayak, R., Neelakantan, A., Ngo, R., Noh, H., Ouyang, L., O'Keefe, C., Pachocki, J., Paino, A., Palermo, J., Pantuliano, A., Parascandolo, G., Parish, J., Parparita, E., Passos, A., Pavlov, M., Peng, A., Perelman, A., de~Avila Belbute~Peres, F., Petrov, M., de~Oliveira~Pinto, H.~P., Michael, Pokorny, Pokrass, M., Pong, V.~H., Powell, T., Power, A., Power, B., Proehl, E., Puri, R., Radford, A., Rae, J., Ramesh, A., Raymond, C., Real, F., Rimbach, K., Ross, C., Rotsted, B., Roussez, H., Ryder, N., Saltarelli, M., Sanders, T., Santurkar, S., Sastry, G., Schmidt, H., Schnurr, D., Schulman, J., Selsam, D., Sheppard, K., Sherbakov, T., Shieh, J., Shoker, S., Shyam, P., Sidor, S., Sigler, E., Simens, M., Sitkin, J., Slama, K., Sohl, I., Sokolowsky, B., Song, Y., Staudacher, N., Such, F.~P., Summers, N., Sutskever, I., Tang, J., Tezak, N., Thompson, M.~B., Tillet, P., Tootoonchian, A., Tseng, E., Tuggle, P., Turley, N., Tworek, J., Uribe, J. F.~C., Vallone, A., Vijayvergiya, A., Voss, C., Wainwright,
  C., Wang, J.~J., Wang, A., Wang, B., Ward, J., Wei, J., Weinmann, C., Welihinda, A., Welinder, P., Weng, J., Weng, L., Wiethoff, M., Willner, D., Winter, C., Wolrich, S., Wong, H., Workman, L., Wu, S., Wu, J., Wu, M., Xiao, K., Xu, T., Yoo, S., Yu, K., Yuan, Q., Zaremba, W., Zellers, R., Zhang, C., Zhang, M., Zhao, S., Zheng, T., Zhuang, J., Zhuk, W., and Zoph, B.
\newblock Gpt-4 technical report, 2024.

\bibitem[Orseau \& Armstrong(2016)Orseau and Armstrong]{safelyinterruptibleagents}
Orseau, L. and Armstrong, S.
\newblock Safely interruptible agents.
\newblock In \emph{Conference on Uncertainty in Artificial Intelligence}, 2016.
\newblock URL \url{https://api.semanticscholar.org/CorpusID:2912679}.

\bibitem[Ouyang et~al.(2022)Ouyang, Wu, Jiang, Almeida, Wainwright, Mishkin, Zhang, Agarwal, Slama, Ray, Schulman, Hilton, Kelton, Miller, Simens, Askell, Welinder, Christiano, Leike, and Lowe]{ouyang2022training}
Ouyang, L., Wu, J., Jiang, X., Almeida, D., Wainwright, C.~L., Mishkin, P., Zhang, C., Agarwal, S., Slama, K., Ray, A., Schulman, J., Hilton, J., Kelton, F., Miller, L., Simens, M., Askell, A., Welinder, P., Christiano, P., Leike, J., and Lowe, R.
\newblock Training language models to follow instructions with human feedback, 2022.

\bibitem[Pan et~al.(2021)Pan, Bhatia, and Steinhardt]{reward_misspecification}
Pan, A., Bhatia, K., and Steinhardt, J.
\newblock The {{Effects}} of {{Reward Misspecification}}: {{Mapping}} and {{Mitigating Misaligned Models}}.
\newblock In \emph{International {{Conference}} on {{Learning Representations}}}, October 2021.

\bibitem[Pang et~al.(2023)Pang, Padmakumar, Sellam, Parikh, and He]{pang_reward_2023}
Pang, R.~Y., Padmakumar, V., Sellam, T., Parikh, A.~P., and He, H.
\newblock Reward {Gaming} in {Conditional} {Text} {Generation}, February 2023.
\newblock URL \url{http://arxiv.org/abs/2211.08714}.
\newblock arXiv:2211.08714 [cs].

\bibitem[Pearl(2009)]{Pearl09}
Pearl, J.
\newblock \emph{Causality}.
\newblock Cambridge University Press, Cambridge, UK, 2 edition, 2009.
\newblock ISBN 978-0-521-89560-6.
\newblock \doi{10.1017/CBO9780511803161}.

\bibitem[Pearl(2019)]{interview_pearl}
Pearl, J.
\newblock Judea pearl: Causal reasoning, counterfactuals, and the path to agi, 2019.
\newblock URL \url{https://youtu.be/pEBI0vF45ic?si=lE_gCwZAFfDU7GG2&t=3869}.

\bibitem[Pepp et~al.(2022)Pepp, Sterken, McKeever, and Michaelson]{Pepp2022-PEPMMB}
Pepp, J., Sterken, R., McKeever, M., and Michaelson, E.
\newblock Manipulative machines (1st edition).
\newblock In Klenk, M. and Jongepier, F. (eds.), \emph{The Philosophy of Online Manipulation}, pp.\  91--107. Routledge, 2022.

\bibitem[Perez et~al.(2022)Perez, Huang, Song, Cai, Ring, Aslanides, Glaese, McAleese, and Irving]{perez2022red}
Perez, E., Huang, S., Song, F., Cai, T., Ring, R., Aslanides, J., Glaese, A., McAleese, N., and Irving, G.
\newblock Red teaming language models with language models, 2022.

\bibitem[Russell(2019)]{russell2019human}
Russell, S.
\newblock \emph{Human Compatible: Artificial Intelligence and the Problem of Control}.
\newblock Penguin Publishing Group, 2019.
\newblock ISBN 9780525558620.
\newblock URL \url{https://books.google.co.uk/books?id=M1eFDwAAQBAJ}.

\bibitem[Russell(2022)]{russell2022}
Russell, S.
\newblock Provably beneficial artificial intelligence.
\newblock In \emph{IUI '22: Proceedings of the 27th International Conference on Intelligent User Interfaces}, 2022.

\bibitem[Russell(2024)]{testimony_russell}
Russell, S.
\newblock Written testimony of {Stuart Russell} before the {U.S. Senate Committee on the Judiciary, Subcommitee on Privacy, Technology, and the Law}, 2024.
\newblock URL \url{https://www.judiciary.senate.gov/imo/media/doc/2023-07-26_-_testimony_-_russell.pdf}.

\bibitem[Saad et~al.(2023)Saad, Patton, Hoffman, A.~Saurous, and Mansinghka]{saad2023sequential}
Saad, F., Patton, B., Hoffman, M.~D., A.~Saurous, R., and Mansinghka, V.
\newblock Sequential {M}onte {C}arlo learning for time series structure discovery.
\newblock In Krause, A., Brunskill, E., Cho, K., Engelhardt, B., Sabato, S., and Scarlett, J. (eds.), \emph{Proceedings of the 40th International Conference on Machine Learning}, volume 202 of \emph{Proceedings of Machine Learning Research}, pp.\  29473--29489. PMLR, 23--29 Jul 2023.
\newblock URL \url{https://proceedings.mlr.press/v202/saad23a.html}.

\bibitem[Sastry et~al.(2024)Sastry, Heim, Belfield, Anderljung, Brundage, Hazell, O'Keefe, Hadfield, Ngo, Pilz, Gor, Bluemke, Shoker, Egan, Trager, Avin, Weller, Bengio, and Coyle]{sastry2024computing}
Sastry, G., Heim, L., Belfield, H., Anderljung, M., Brundage, M., Hazell, J., O'Keefe, C., Hadfield, G.~K., Ngo, R., Pilz, K., Gor, G., Bluemke, E., Shoker, S., Egan, J., Trager, R.~F., Avin, S., Weller, A., Bengio, Y., and Coyle, D.
\newblock Computing power and the governance of artificial intelligence, 2024.

\bibitem[Schlaginhaufen \& Kamgarpour(2023)Schlaginhaufen and Kamgarpour]{schlaginhaufen2023identifiability}
Schlaginhaufen, A. and Kamgarpour, M.
\newblock Identifiability and generalizability in constrained inverse reinforcement learning, 2023.

\bibitem[Schuett et~al.(2023)Schuett, Dreksler, Anderljung, McCaffary, Heim, Bluemke, and Garfinkel]{schuett2023best}
Schuett, J., Dreksler, N., Anderljung, M., McCaffary, D., Heim, L., Bluemke, E., and Garfinkel, B.
\newblock Towards best practices in agi safety and governance: A survey of expert opinion, 2023.

\bibitem[Seligman et~al.(2023)Seligman, Schubert, and Kumar]{seligman2023formal}
Seligman, E., Schubert, T., and Kumar, M. A.~K.
\newblock \emph{Formal verification: an essential toolkit for modern VLSI design}.
\newblock Elsevier, 2023.

\bibitem[Seshia(2015)]{seshia-pieee15}
Seshia, S.~A.
\newblock Combining induction, deduction, and structure for verification and synthesis.
\newblock \emph{Proceedings of the {IEEE}}, 103\penalty0 (11):\penalty0 2036--2051, 2015.

\bibitem[Seshia et~al.(2016)Seshia, Sadigh, and Sastry]{seshia-arxiv16}
Seshia, S.~A., Sadigh, D., and Sastry, S.~S.
\newblock {Towards Verified Artificial Intelligence}.
\newblock \emph{ArXiv e-prints}, July 2016.

\bibitem[Seshia et~al.(2018)Seshia, Desai, Dreossi, Fremont, Ghosh, Kim, Shivakumar, Vazquez-Chanlatte, and Yue]{seshia-atva18}
Seshia, S.~A., Desai, A., Dreossi, T., Fremont, D., Ghosh, S., Kim, E., Shivakumar, S., Vazquez-Chanlatte, M., and Yue, X.
\newblock Formal specification for deep neural networks.
\newblock In \emph{Proceedings of the International Symposium on Automated Technology for Verification and Analysis (ATVA)}, pp.\  20--34, October 2018.

\bibitem[Seshia et~al.(2022)Seshia, Sadigh, and Sastry]{seshia-cacm22a}
Seshia, S.~A., Sadigh, D., and Sastry, S.~S.
\newblock Toward verified artificial intelligence.
\newblock \emph{Communications of the {ACM}}, 65\penalty0 (7):\penalty0 46--55, 2022.

\bibitem[Settle(2018)]{Settle_2018}
Settle, J.~E.
\newblock \emph{Frenemies: How Social Media Polarizes America}.
\newblock Cambridge University Press, 2018.

\bibitem[Shah et~al.(2022)Shah, Varma, Kumar, Phuong, Krakovna, Uesato, and Kenton]{shah2022goal}
Shah, R., Varma, V., Kumar, R., Phuong, M., Krakovna, V., Uesato, J., and Kenton, Z.
\newblock Goal misgeneralization: Why correct specifications aren't enough for correct goals, 2022.

\bibitem[Skalse \& Abate(2023{\natexlab{a}})Skalse and Abate]{pmlr-v216-skalse23a}
Skalse, J. and Abate, A.
\newblock On the limitations of {M}arkovian rewards to express multi-objective, risk-sensitive, and modal tasks.
\newblock In Evans, R.~J. and Shpitser, I. (eds.), \emph{Proceedings of the Thirty-Ninth Conference on Uncertainty in Artificial Intelligence}, volume 216 of \emph{Proceedings of Machine Learning Research}, pp.\  1974--1984. PMLR, 31 Jul--04 Aug 2023{\natexlab{a}}.
\newblock URL \url{https://proceedings.mlr.press/v216/skalse23a.html}.

\bibitem[Skalse \& Abate(2023{\natexlab{b}})Skalse and Abate]{skalse2023misspecification}
Skalse, J. and Abate, A.
\newblock Misspecification in inverse reinforcement learning, 2023{\natexlab{b}}.

\bibitem[Skalse \& Abate(2024)Skalse and Abate]{skalse2024quantifying}
Skalse, J. and Abate, A.
\newblock Quantifying the sensitivity of inverse reinforcement learning to misspecification, 2024.

\bibitem[Skalse et~al.(2024)Skalse, Farnik, Motwani, Jenner, Gleave, and Abate]{skalse2024starc}
Skalse, J., Farnik, L., Motwani, S.~R., Jenner, E., Gleave, A., and Abate, A.
\newblock Starc: A general framework for quantifying differences between reward functions, 2024.

\bibitem[Skalse et~al.(2022)Skalse, Howe, Krasheninnikov, and Krueger]{skalse_defining_2022}
Skalse, J. M.~V., Howe, N. H.~R., Krasheninnikov, D., and Krueger, D.
\newblock Defining and {{Characterizing Reward Gaming}}.
\newblock In \emph{Advances in {{Neural Information Processing Systems}}}, May 2022.

\bibitem[Skalse et~al.(2023)Skalse, {Farrugia-Roberts}, Russell, Abate, and Gleave]{skalse2022invariance}
Skalse, J. M.~V., {Farrugia-Roberts}, M., Russell, S., Abate, A., and Gleave, A.
\newblock Invariance in policy optimisation and partial identifiability in reward learning.
\newblock In \emph{International {{Conference}} on {{Machine Learning}}}, pp.\  32033--32058. {PMLR}, 2023.

\bibitem[Soares et~al.(2015)Soares, Fallenstein, Armstrong, and Yudkowsky]{corrigibility}
Soares, N., Fallenstein, B., Armstrong, S., and Yudkowsky, E.
\newblock Corrigibility.
\newblock In Walsh, T. (ed.), \emph{Artificial Intelligence and Ethics, Papers from the 2015 {AAAI} Workshop, Austin, Texas, USA, January 25, 2015}, volume {WS-15-02} of \emph{{AAAI} Technical Report}. {AAAI} Press, 2015.
\newblock URL \url{http://aaai.org/ocs/index.php/WS/AAAIW15/paper/view/10124}.

\bibitem[Stray(2021)]{stray2021designing}
Stray, J.
\newblock Designing recommender systems to depolarize, 2021.

\bibitem[Stuhlm{\"u}ller et~al.(2015)Stuhlm{\"u}ller, Hawkins, Siddharth, and Goodman]{stuhlmuller2015coarse}
Stuhlm{\"u}ller, A., Hawkins, R.~X., Siddharth, N., and Goodman, N.~D.
\newblock Coarse-to-fine sequential {M}onte {C}arlo for probabilistic programs.
\newblock \emph{arXiv preprint arXiv:1509.02962}, 2015.

\bibitem[Subramani et~al.(2024)Subramani, Williams, Heitmann, Holm, Griffin, and Skalse]{subramani2024expressivity}
Subramani, R., Williams, M., Heitmann, M., Holm, H., Griffin, C., and Skalse, J.
\newblock On the expressivity of objective-specification formalisms in reinforcement learning, 2024.

\bibitem[Szegedy(2020)]{szegedy2020promising}
Szegedy, C.
\newblock A promising path towards autoformalization and general artificial intelligence.
\newblock In \emph{Intelligent Computer Mathematics: 13th International Conference, CICM 2020, Bertinoro, Italy, July 26--31, 2020, Proceedings 13}, pp.\  3--20. Springer, 2020.

\bibitem[Tabuada(2009)]{tabuada2009verification}
Tabuada, P.
\newblock \emph{Verification and control of hybrid systems: a symbolic approach}.
\newblock Springer Science \& Business Media, 2009.

\bibitem[Tang et~al.(2024)Tang, Key, and Ellis]{tang2024worldcoder}
Tang, H., Key, D., and Ellis, K.
\newblock Worldcoder, a model-based llm agent: Building world models by writing code and interacting with the environment.
\newblock \emph{arXiv preprint arXiv:2402.12275}, 2024.

\bibitem[Tegmark(2018)]{tegmark2018life}
Tegmark, M.
\newblock \emph{Life 3.0: Being human in the age of artificial intelligence}.
\newblock Vintage, 2018.

\bibitem[Tegmark \& Omohundro(2023)Tegmark and Omohundro]{tegmark2023provably}
Tegmark, M. and Omohundro, S.
\newblock Provably safe systems: the only path to controllable agi, 2023.

\bibitem[Trinh et~al.(2024)Trinh, Wu, Le, He, and Luong]{Trinh2024}
Trinh, T.~H., Wu, Y., Le, Q.~V., He, H., and Luong, T.
\newblock Solving olympiad geometry without human demonstrations.
\newblock \emph{Nature}, 625\penalty0 (7995):\penalty0 476--482, Jan 2024.
\newblock ISSN 1476-4687.
\newblock \doi{10.1038/s41586-023-06747-5}.
\newblock URL \url{https://doi.org/10.1038/s41586-023-06747-5}.

\bibitem[Turchin(2021)]{turchin2021multilevel}
Turchin, A.
\newblock Catching treacherous turn: A model of the multilevel ai boxing, 05 2021.

\bibitem[Turing(1951)]{turing1951}
Turing, A.
\newblock Intelligent machinery, a heretical theory.
\newblock 1951.

\bibitem[van Merwijk et~al.(2022)van Merwijk, Carey, and Everitt]{vanmerwijk2022complete}
van Merwijk, C., Carey, R., and Everitt, T.
\newblock A complete criterion for value of information in soluble influence diagrams, 2022.

\bibitem[Vazquez{-}Chanlatte \& Seshia(2020)Vazquez{-}Chanlatte and Seshia]{vazquez-cav20}
Vazquez{-}Chanlatte, M. and Seshia, S.~A.
\newblock Maximum causal entropy specification inference from demonstrations.
\newblock In \emph{32nd International Conference on Computer Aided Verification (CAV)}, July 2020.

\bibitem[Vazquez{-}Chanlatte et~al.(2018)Vazquez{-}Chanlatte, Jha, Tiwari, Ho, and Seshia]{vazquez-neurips18}
Vazquez{-}Chanlatte, M., Jha, S., Tiwari, A., Ho, M.~K., and Seshia, S.~A.
\newblock Learning task specifications from demonstrations.
\newblock In \emph{Advances in Neural Information Processing Systems 31: Annual Conference on Neural Information Processing Systems (NeurIPS)}, pp.\  5372--5382, December 2018.

\bibitem[Viano et~al.(2021)Viano, Huang, Kamalaruban, Weller, and Cevher]{viano2021robust}
Viano, L., Huang, Y.-T., Kamalaruban, P., Weller, A., and Cevher, V.
\newblock Robust inverse reinforcement learning under transition dynamics mismatch.
\newblock In Beygelzimer, A., Dauphin, Y., Liang, P., and Vaughan, J.~W. (eds.), \emph{Advances in Neural Information Processing Systems}, 2021.
\newblock URL \url{https://openreview.net/forum?id=t8HduwpoQQv}.

\bibitem[Vinge(1993)]{vinge1993}
Vinge, V.
\newblock The coming technological singularity: How to survive in the post-human era.
\newblock \emph{Whole Earth Review}, 1993.

\bibitem[Wang et~al.(2022)Wang, Zhang, and Zhu]{fairness_review_2}
Wang, X., Zhang, Y., and Zhu, R.
\newblock A brief review on algorithmic fairness.
\newblock \emph{Management System Engineering}, 1\penalty0 (1):\penalty0 7, Nov 2022.
\newblock ISSN 2731-5843.
\newblock \doi{10.1007/s44176-022-00006-z}.
\newblock URL \url{https://doi.org/10.1007/s44176-022-00006-z}.

\bibitem[Watanabe(2009)]{Watanabe_2009}
Watanabe, S.
\newblock \emph{Algebraic Geometry and Statistical Learning Theory}.
\newblock Cambridge Monographs on Applied and Computational Mathematics. Cambridge University Press, 2009.

\bibitem[Watanabe(2018)]{Watanabe_2018}
Watanabe, S.
\newblock \emph{Mathematical Theory of Bayesian Statistics}.
\newblock Chapman and Hall/CRC, 2018.

\bibitem[Wiener(1950)]{wiener1950}
Wiener, N.
\newblock \emph{The Human Use of Human Beings}.
\newblock Eyre \& Spottiswoode, 1950.

\bibitem[Wing(2021)]{wing2021}
Wing, J.~M.
\newblock Trustworthy {AI}.
\newblock \emph{Communications of the ACM}, 64\penalty0 (10):\penalty0 64–71, September 2021.
\newblock ISSN 1557-7317.
\newblock \doi{10.1145/3448248}.
\newblock URL \url{http://dx.doi.org/10.1145/3448248}.

\bibitem[Wong et~al.(2023{\natexlab{a}})Wong, Grand, Lew, Goodman, Mansinghka, Andreas, and Tenenbaum]{wong2023word}
Wong, L., Grand, G., Lew, A.~K., Goodman, N.~D., Mansinghka, V.~K., Andreas, J., and Tenenbaum, J.~B.
\newblock From word models to world models: Translating from natural language to the probabilistic language of thought, 2023{\natexlab{a}}.

\bibitem[Wong et~al.(2023{\natexlab{b}})Wong, Mao, Sharma, Siegel, Feng, Korneev, Tenenbaum, and Andreas]{wong2023learning}
Wong, L., Mao, J., Sharma, P., Siegel, Z.~S., Feng, J., Korneev, N., Tenenbaum, J.~B., and Andreas, J.
\newblock Learning adaptive planning representations with natural language guidance.
\newblock \emph{arXiv preprint arXiv:2312.08566}, 2023{\natexlab{b}}.

\bibitem[Wu et~al.(2022)Wu, Jiang, Li, Rabe, Staats, Jamnik, and Szegedy]{wu2022autoformalization}
Wu, Y., Jiang, A.~Q., Li, W., Rabe, M., Staats, C., Jamnik, M., and Szegedy, C.
\newblock Autoformalization with large language models.
\newblock \emph{Advances in Neural Information Processing Systems}, 35:\penalty0 32353--32368, 2022.

\bibitem[Yalcinkaya et~al.(2023)Yalcinkaya, Torfah, Fremont, and Seshia]{yalcinkaya-rv23}
Yalcinkaya, B., Torfah, H., Fremont, D.~J., and Seshia, S.~A.
\newblock Compositional simulation-based analysis of {AI}-based autonomous systems for markovian specifications.
\newblock In \emph{Runtime Verification - 23rd International Conference (RV)}, volume 14245 of \emph{Lecture Notes in Computer Science}, pp.\  191--212. Springer, 2023.

\bibitem[Yudkowsky(2001)]{yudkowsky2001}
Yudkowsky, E.
\newblock Creating friendly ai 1.0: The analysis and design of benevolent goal architectures.
\newblock \emph{The Singularity Institute}, 2001.

\bibitem[Zhang(2022)]{zhang2022viper}
Zhang, D.
\newblock Adding native support for havoc in viper.
\newblock \emph{Proceedings of IEEE International Conference on Robotics and Automation}, 2022.

\bibitem[Zhang et~al.(2021)Zhang, Tino, Leonardis, and Tang]{zhang2021interpretability}
Zhang, Y., Tino, P., Leonardis, A., and Tang, K.
\newblock A survey on neural network interpretability.
\newblock \emph{IEEE Transactions on Emerging Topics in Computational Intelligence}, 5\penalty0 (5):\penalty0 726–742, October 2021.
\newblock ISSN 2471-285X.
\newblock \doi{10.1109/tetci.2021.3100641}.
\newblock URL \url{http://dx.doi.org/10.1109/TETCI.2021.3100641}.

\bibitem[Zhao et~al.(2024)Zhao, Brekelmans, Makhzani, and Grosse]{zhao2024probabilistic}
Zhao, S., Brekelmans, R., Makhzani, A., and Grosse, R.
\newblock Probabilistic inference in language models via twisted sequential monte carlo.
\newblock \emph{arXiv preprint arXiv:2404.17546}, 2024.

\bibitem[Zhuang \& {Hadfield-Menell}(2020)Zhuang and {Hadfield-Menell}]{subset_features}
Zhuang, S. and {Hadfield-Menell}, D.
\newblock Consequences of misaligned {{AI}}.
\newblock In \emph{Proceedings of the 34th {{International Conference}} on {{Neural Information Processing Systems}}}, {{NIPS}}'20, pp.\  15763--15773, {Red Hook, NY, USA}, December 2020. {Curran Associates Inc.}
\newblock ISBN 978-1-71382-954-6.

\bibitem[Ziegler et~al.(2022)Ziegler, Nix, Chan, Bauman, Schmidt-Nielsen, Lin, Scherlis, Nabeshima, Weinstein-Raun, de~Haas, Shlegeris, and Thomas]{ziegler2022adversarial}
Ziegler, D.~M., Nix, S., Chan, L., Bauman, T., Schmidt-Nielsen, P., Lin, T., Scherlis, A., Nabeshima, N., Weinstein-Raun, B., de~Haas, D., Shlegeris, B., and Thomas, N.
\newblock Adversarial training for high-stakes reliability, 2022.

\end{thebibliography}
\bibliographystyle{icml2024}

\newpage
\appendix

\section{A Definition of Harm}\label{appendix:harm}

Many proposed safety specifications require recognizing when harm occurs. To better understand how to represent harm in a safety specification, it's helpful to consider how this concept can be formalized using tools such as causal models.

Before we can define harm, it’s worth first reviewing the basics of causal models: we can view a causal model as a directed acyclic graph (DAG), where nodes represent variables and edges represent causal relationships between those variables. In the DAG, we take the variables at the top to be \textit{exogenous} (their values are determined by factors outside the scope of the mode), while all other variables are \textit{endogenous} and have associated equations that determine their values as a function of the values of their parents (replacing the conditional probability tables that characterize a Bayesian network by equations).

The equations in the causal model represent how things work in the absence of interventions. They describe the causal relationships between variables and determine the outcomes based on the values of the parent variables. However, the causal model also allows for interventions. If an intervention is made, such as setting variable $X$ to a specific value (e.g., $X=3$), a new causal model is created. In this new model, everything upstream of the intervention remains the same, while everything downstream is changed according to the equations and the intervention.

While the actual definition of causation is somewhat more complicated, for our purposes, it suffices to say that $A$ rather than $A'$ is a cause of $B$ rather than $B'$ if, had $A'$ happened instead of $A$, $B'$ would have happened rather than $B$, where $A$ and $A'$ are different settings of the same variables, as are $B$ and $B'$.  So we can say, for example, that $X=2$ rather than $X=3$ is a cause of $Y=1$ rather than $Y=0$ if, had $X$ been set to 3 (instead of taking its actual value of 2), then $Y$ would have been 1 rather than its actual value of 0.

In order to capture harm, we need to extend the standard causal model with two additional features:

\begin{itemize}
\item \textit{Utility on outcomes}: The causal model includes a special endogenous variable called "outcome," which represents the possible outcomes of the scenario being modeled. Each value of the outcome variable is associated with a specific utility value, normalized between 0 and 1. For example, if outcome X has a utility of 1, it means that X is the most desirable outcome in the context of the model.   (The assumption that there is a single outcome, as opposed to a set of outcomes, and that its value is normalized to being in [0,1], can both be made without loss of generality.  We can, for example, always “package up” a set of outcomes into a single variable.)

\item \textit{Default utility}: The causal model also includes a default utility value, which is context-dependent. This default utility represents the expected or "normal" utility in the given scenario, and it is used as a reference point to determine whether an event causes harm.
\end{itemize}

With this background, we are in a position to talk about what it means for an event to cause harm in an extended causal model:

If an event caused a specific outcome whose utility is worse than a default utility (which is given as part of the model) and there’s an alternative event which would have led to a utility with a better outcome, then harm was caused. A little more formally, an event $X=x$ causes harm in a causal setting if there exist values $x'$ and outcomes $o,o'$ such that (i) the utility of the actual outcome $o$ is less than the default utility, and (ii) the event $X=x$ rather than $X=x'$ causes outcome $o$ rather than $o'$, where the utility of $o$ is less than the utility of $o'$.

We find this definition of harm an appealing starting point, which could then be defined and bound in a safety specification and verified against a world model. However, we do think this is just a starting point; implementing this in a complex world model to meaningfully cover the utilities, the downstream implications of an event, etc. are, to put it mildly, a challenge.

\section{A Formal Definition of GS AI}\label{appendix:formal_definition}

In this Appendix, we provide a more formal version of Definition~\ref{def:GSAI}.

\begin{definition}\label{def:GSAI_formal}
A \emph{Guaranteed Safe AI system} is one that is equipped with a quantitative safety guarantee, that is produced by a world-model, a safety specification, and a verifier. Formally, this can be described as follows:
\begin{itemize}
    \item Let $A$ be a set of \emph{actions}, $S$ be a set of \emph{world states}, and $O$ be a set of \emph{observation}. A \emph{policy} $\pi$ is a function $\pi : (O \times A)^\star \times O \to \Delta(A)$, which returns a distribution of actions for each sequence of past observations and actions. Let $\Pi$ be the set of all policies.
    \item A \emph{world model} $m$ is a function $m : S \times A \to \mathcal{P}(\Delta(O \times S))$, which returns a set of probability distributions of world states and observations for each past world state and action. Let $M$ be the set of all world models.\footnote{Note that this includes the case where the world model always outputs a precise probability distribution, i.e.\ functions $m : S \times A \to \Delta(O \times S)$. However, we also include the case where the world model outputs a credal set, for example.}
    \item A \emph{safety specification} $\psi$ is a function $\phi : \Pi \times M \to [0,1]$, which specifies a value between $0$ and $1$ for each policy and world model. 
    \item A \emph{verifier} $V_\psi$ for a specification $\psi$ is an algorithm $V_\psi : \Pi \times M \to \mathcal{P}([0,1])$ that computes or estimates the value of $\psi(\pi, m)$.\footnote{In general, we permit $V_\psi$ to be an anytime algorithm which never terminates, but instead is productive of a sequence of increasingly accurate estimates. In these cases $V_\psi$ also takes as input a time parameter $t \in \mathbb{N}$, and should satisfy that $V_\psi(\pi,m,t) \subseteq V_\psi(\pi,m,t')$ whenever $t \leq t'$. We can also allow $V_\psi$ to be probabilistic, in which case it also takes a random vector as input.} 
\end{itemize}
A GS AI system is thus a tuple $\langle \pi, m, \psi, V_\psi \rangle$, where $\pi$ is a policy such that $V_\psi(\pi, m)$ lies in some permissible set.
    
\end{definition}

\end{document}